\documentclass[11pt]{article}

\usepackage[preprint]{acl}

\usepackage{times}
\usepackage{latexsym}
\usepackage{amsmath} 
\usepackage{amsfonts} 
\usepackage{algorithm} 
\usepackage{algorithmic} 

\usepackage{amssymb}
\usepackage{mathtools}
\usepackage{amsthm}
\usepackage{booktabs}
\usepackage{multirow} 
\usepackage[T1]{fontenc}

\usepackage[utf8]{inputenc}

\usepackage{microtype}

\usepackage{inconsolata}

\usepackage{graphicx}
\usepackage{subcaption}  
\usepackage{authblk}

\title{Steering Vision-Language Models with Joint Sparse Autoencoders}

\author[1]{Huizhen Shu}
\author[2]{Xuying Li}
\author[3]{Hongxu Lin}
\author[4]{Wenjie Sun}
\author[1]{Hui Li}

\affil[1]{yunshanai, \texttt{\{shuhuizhen, lihui\}@yunshanai.org.cn}}
\affil[2]{HKUST(Guangzhou), \texttt{xli118@connect.hkust-gz.edu.cn}}
\affil[3]{Zidongtaichu, \texttt{linhongxu@taichu.ai}}
\affil[4]{University of British Columbia, \texttt{wenjie1835@gmail.com}}

\begin{document}
\maketitle
\begin{abstract}
Sparse Autoencoders (SAEs) have shown promise for analyzing language models, but applying them to vision-language models (VLMs) often yields representations that are difficult to use as controllable cross-modal steering directions. We introduce the Joint Sparse Autoencoder (\textbf{JSAE}), which uses an explicit alignment constraint to jointly factorize \emph{sequence-pooled} vision and language activations into shared, interpretable image/caption-level features. Applied to LLaVA, JSAE recovers cross-modal features for recognizable concepts (e.g., \textit{food}, \textit{animals}). Through bidirectional interventions (additive steering and suppression), we observe a layer-dependent asymmetry under our protocol: additive steering peaks at mid-to-late (pre-output) layers and weakens at both ends, whereas suppression scores remain within a comparable range across all probed layers (within statistical noise). Experiments on three VLMs---LLaVA-v1.6-Mistral-7B, Llama3-LLaVA-8B, and the MoE-based Qwen3-VL-30B---show related layer-localized effects across architectures. Together, these results suggest that explicitly aligned sparse representations support more controllable intervention-based analysis of multimodal features, within an identifiable layer range, than the unconstrained alternatives tested here.
\end{abstract}

\section{Introduction}

Vision-Language Models (VLMs) have become an important class of modern multimodal systems, supporting image captioning, visual reasoning, and embodied decision-making by aligning visual inputs with linguistic concepts~\citep{li2023blip2bootstrappinglanguageimagepretraining,driess2023palmeembodiedmultimodallanguage}. A key ingredient behind these capabilities is \emph{multimodal alignment}: the ability to associate visual patterns with semantic abstractions expressed in language. Despite its importance, the internal mechanisms by which VLMs support this alignment remain incompletely understood. In particular, it is unclear \textit{where} in the network this alignment emerges, \textit{how} vision and language representations are coordinated at the feature level, and whether these learned representations have an intervention-relevant role in text generation rather than merely reflecting statistical correlations.

Sparse Autoencoders (SAEs) decompose polysemantic hidden states into interpretable latent directions in large language models~\citep{bricken2023monosemantic,cunningham2023sparseautoencodershighlyinterpretable,marks2025sparsefeaturecircuitsdiscovering}. Extending them to VLMs is non-trivial: recent works such as VL-SAE~\citep{shen2025vlsaeinterpretingenhancingvisionlanguage} share a single dictionary across modalities or align sparse codes via co-activation statistics on the raw stream. Our setting instead requires \emph{paired-code directional alignment} between two modality-specific dictionaries that can serve as cross-modal steering directions---a property not directly provided by these formulations.

In this work, we address these limitations by introducing a unified framework for discovering and examining multimodal alignment features. Our key insight is that effective interpretability in VLMs benefits from \emph{jointly trained} sparse representations with explicit constraints that encourage cross-modal coordination. To this end, we propose a pipeline that combines: (i) layer-wise localization to identify stages where alignment is most separable; (ii) \textbf{Joint Sparse Autoencoders (JSAE)} that simultaneously factorize \emph{mean-pooled} image- and caption-level activations into a shared latent space via a cosine-similarity penalty applied to paired sparse codes; (iii) directional clustering to group and select the most consistent semantic features; and (iv) bidirectional interventions (additive steering and feature suppression) to probe the functional role of these features. We additionally provide a geometric design rationale for why JSAE's alignment loss is expected to induce modality-paired decoder columns, and why text-side directions are expected to be useful steering vectors at mid-to-late layers but to weaken near the output boundary; these expectations are then tested empirically (Section~\ref{sec:results}).

Applying this pipeline to LLaVA-v1.6-Mistral-7B~\citep{liu2024improvedbaselinesvisualinstruction}, we identify sparse features associated with semantic concepts such as \textit{food} and \textit{animals}. Our intervention analyses suggest a layer-dependent asymmetry between the two intervention modes under our protocol: additive steering peaks at mid-to-late (pre-output) layers and weakens at both ends of the depth range, whereas feature suppression remains within a comparable score range across all probed layers. We further test whether similar patterns appear in dense and MoE-based VLMs (Qwen3-VL-30B~\citep{bai2025qwen3vltechnicalreport} and Llama3-LLaVA-8B~\citep{li2024llavanext-strong}).

\begin{figure*}[t]
\centering
\includegraphics[width=\textwidth]{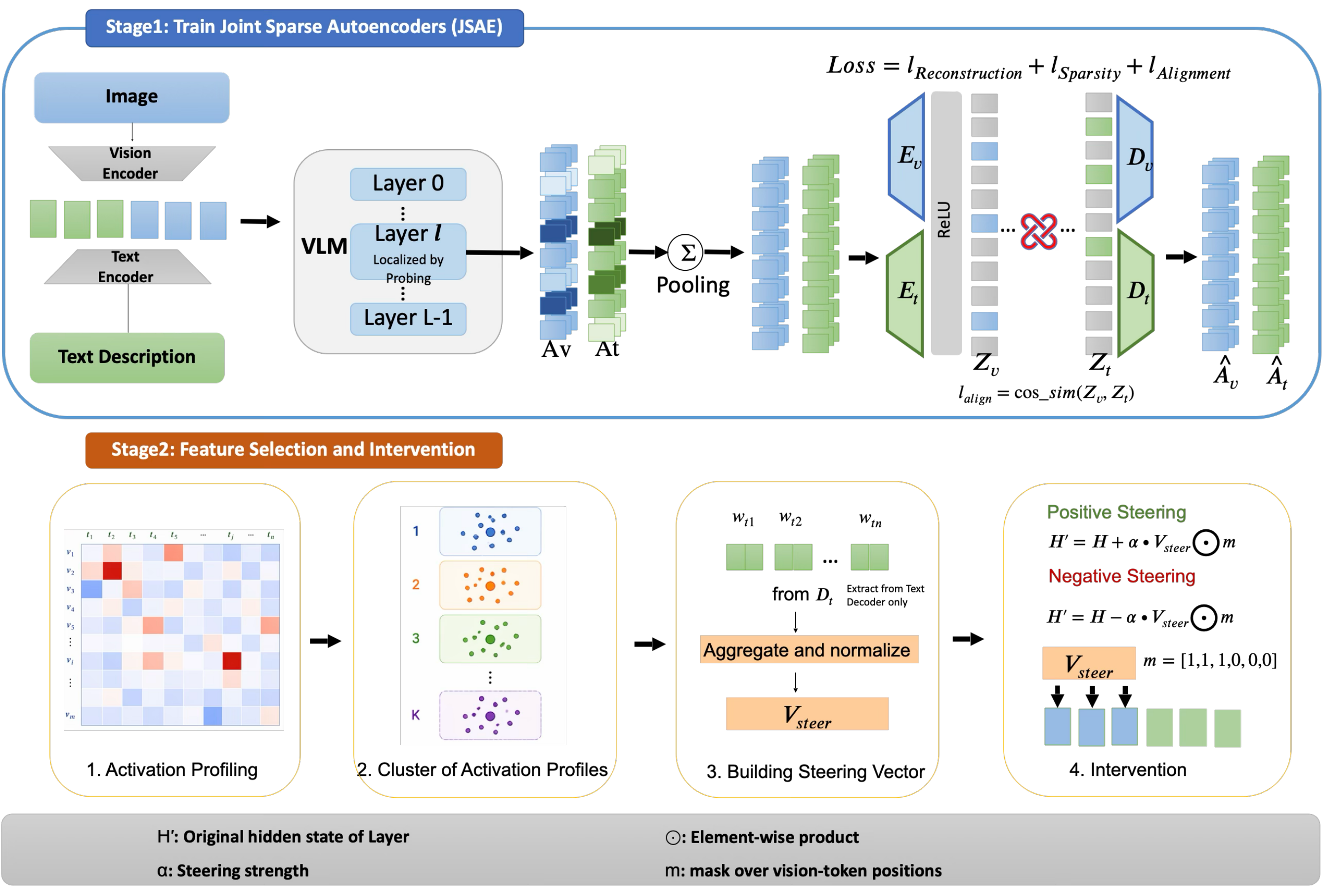} 
\caption{Overview of the JSAE Framework. \textbf{Stage 1 (Training):} Vision and text activations are factorized into a shared overcomplete latent space, with a cosine-similarity constraint ($l_{align}$) aligning their decoder directions. \textbf{Stage 2 (Intervention):} Aligned semantic directions are extracted from the Text Decoder and additively injected onto image-token positions (mask $m$) during the forward pass, steering visual processing and subsequent generation.}
\label{fig:main_architecture}
\end{figure*}
\section{Related Work}

\subsection{Multimodal Alignment}

Multimodal alignment connects visual patterns with linguistic abstractions in a shared representational space. Contrastive pretraining approaches like CLIP/ALIGN~\citep{radford2021learningtransferablevisualmodels,6} learn transferable image-text representations, while generative VLMs (Flamingo~\citep{7}, BLIP-2~\citep{li2023blip2bootstrappinglanguageimagepretraining}, InstructBLIP~\citep{8}, MiniGPT-4~\citep{10}, LLaVA~\citep{liu2024improvedbaselinesvisualinstruction}, Qwen-VL~\citep{9}, PaLM-E~\citep{11}) bridge vision encoders and language models via cross-attention, query transformers, or projection modules. Their alignment is typically optimized at the input-output level rather than characterized at the feature level. \citet{venhoff2025visualrepresentationsmaplanguage} examine how visual information enters the language feature space and identify a layer-wise progression toward language-aligned features. We follow this mechanistic direction but learn sparse cross-modal features directly from VLM activations and use them for intervention-based control of generation.

\subsection{Sparse Autoencoders for Interpretability}

Sparse Autoencoders (SAEs) have become a prominent tool for decomposing dense neural activations into sparse and more interpretable latent features. Motivated by the superposition hypothesis and the problem of polysemanticity~\citep{elhage2022toymodelssuperposition}, SAEs learn overcomplete dictionaries that reconstruct model activations while encouraging sparse feature usage. Prior work has shown that SAEs can recover interpretable and often more monosemantic features in language models~\citep{bricken2023monosemantic,cunningham2023sparseautoencodershighlyinterpretable}, scale to large models and large dictionaries~\citep{templeton2024scaling,1}, and serve as units for causal or circuit-level analysis~\citep{marks2025sparsefeaturecircuitsdiscovering}. Recent architectural variants such as Gated SAE, TopK SAE, JumpReLU SAE, BatchTopK SAE and LocA-SAE further improve the reconstruction-sparsity trade-off or address limitations such as shrinkage and feature splitting~\citep{gao2025weightsparsetransformersinterpretablecircuits,2,3,4,5}.

SAEs have also recently been extended to multimodal settings. VL-SAE~\citep{shen2025vlsaeinterpretingenhancingvisionlanguage} uses a shared sparse code dictionary across modalities and aligns vision/text activations via co-activation statistics on the residual stream, while Group-Sparse Autoencoders~\citep{kaushik2026decomposingmultimodalembeddingspaces} address modality-splitting by imposing structured sparsity across multimodal dictionaries. Our work is closest to these multimodal SAE approaches, but differs in two specific ways: (i) JSAE maintains \emph{modality-specific} encoder-decoder pairs with paired-code cosine alignment, so each modality keeps its own dictionary directions while being directionally bound to its cross-modal counterpart; (ii) we evaluate whether the resulting text-side dictionary directions can be applied as \emph{causal steering vectors on the visual stream}, which is a specific intervention regime not directly addressed by prior multimodal SAE formulations.

\subsection{Controllable Generation via Activation Engineering}

Controllable generation is commonly achieved through prompt engineering, fine-tuning, RLHF, or parameter-efficient adaptation~\citep{liu2023visualinstructiontuning,gao2023llamaadapterv2parameterefficientvisual}, which control behavior externally rather than by identifying responsible internal representations. Activation-level methods instead manipulate hidden states at inference time: Activation Addition builds steering vectors from contrastive activation differences~\citep{turner2024steeringlanguagemodelsactivation}, Representation Engineering manipulates population-level representations~\citep{12}, and causal tracing or model editing localizes internal computations~\citep{13,14}. Most prior steering work is unimodal; extending it to VLMs is nontrivial because vision/text activations may occupy partially misaligned geometries and direct image-token interventions can drift off-distribution. JSAE addresses this by learning aligned sparse directions across modalities; additive steering tests whether a feature direction can induce target-concept behavior, while suppression tests whether reducing it weakens the corresponding concept.

\section{Method}

\subsection{Joint Sparse Autoencoder for Multimodal Alignment}
\label{sec:method_jsae}

To decompose multimodal representations into interpretable cross-modal features, we train a Joint Sparse Autoencoder (JSAE) that
simultaneously factorizes vision and text activations into a shared latent space. Since a VLM processes a mixed sequence of
varying-length image patch tokens ($N$ tokens) and text tokens ($M$ tokens), we first decouple them using modality-specific binary
masks. To stably capture cross-modal global semantics without being biased by sequence lengths, we apply mean-pooling over the
sequence dimensions of the target layer's hidden states to compute macroscopic representations:
\begin{equation}
\mathbf{A}_v = \frac{1}{N}\sum_{i=1}^{N} \mathbf{h}_{v,i}, \quad \mathbf{A}_t = \frac{1}{M}\sum_{j=1}^{M} \mathbf{h}_{t,j}
\end{equation}
Given these modality-pooled activations $\mathbf{A}_v, \mathbf{A}_t \in \mathbb{R}^{d}$, JSAE learns separate encoder-decoder pairs
$(\mathbf{E}_v, \mathbf{D}_v)$ and $(\mathbf{E}_t, \mathbf{D}_t)$ that map to an overcomplete sparse latent space $\mathbb{R}^{k}$ ($k
 \gg d$).\footnote{Unlike token-level SAE work that decomposes individual residual-stream states, JSAE here operates on \emph{sequence-pooled} representations; its features are thus image/caption-level macro-concepts rather than token-level monosemantic units. We discuss this trade-off in ``Limitations and Future Work''.}

\paragraph{Architecture \& Objective.}
Each modality is encoded independently: $\mathbf{z}_v = \text{ReLU}(\mathbf{E}_v(\mathbf{A}_v))$ and $\mathbf{z}_t =
\text{ReLU}(\mathbf{E}_t(\mathbf{A}_t))$, then reconstructed via their respective decoders: $\hat{\mathbf{A}}_v =
\mathbf{D}_v(\mathbf{z}_v)$ and $\hat{\mathbf{A}}_t = \mathbf{D}_t(\mathbf{z}_t)$. The JSAE is optimized using a composite loss that
balances reconstruction fidelity, sparsity, and explicit cross-modal alignment:
\begin{equation}
\mathcal{L}_{\text{total}} = \lambda_r \mathcal{L}_{\text{recon}} + \lambda_s \mathcal{L}_{\text{sparse}} + \lambda_a
\mathcal{L}_{\text{align}}
\label{eq:jsae_loss}
\end{equation}
where:
\begin{itemize}
    \item \textbf{Reconstruction:} $\mathcal{L}_{\text{recon}} = \frac{1}{2}(\|\hat{\mathbf{A}}_v - \mathbf{A}_v\|^2 +
\|\hat{\mathbf{A}}_t - \mathbf{A}_t\|^2)$ preserves the original activation information.
    \item \textbf{Sparsity:} $\mathcal{L}_{\text{sparse}} = \frac{1}{2}(\|\mathbf{z}_v\|_1 + \|\mathbf{z}_t\|_1)$ drives activations
toward zero (soft $L_1$ sparsity).
    \item \textbf{Alignment:} $\mathcal{L}_{\text{align}} = \frac{1}{B}\sum_{b=1}^{B}(1 - \text{cosine\_sim}(\mathbf{z}_v^{(b)},
\mathbf{z}_t^{(b)}))$ encourages paired macro-level vision-text representations to align directionally. For sparse non-negative codes,
this penalty biases the encoders toward selecting the same subset of latent dimensions across modalities.
\end{itemize}
This training regime encourages overlapping active latent subsets for semantically related vision-text pairs, providing a practical
basis for the intervention analyses below.

\paragraph{Design Rationale.}
The cosine-alignment term provides a geometric intuition for why JSAE's text-decoder columns are expected to serve as cross-modal steering directions; we state it here informally as a design rationale rather than a formal result, and all of its consequences are tested empirically in Section~\ref{sec:results}. Because $\mathbf{z}_v, \mathbf{z}_t \in \mathbb{R}^{k}_{\geq 0}$ after ReLU,
\begin{equation}
\cos(\mathbf{z}_v, \mathbf{z}_t) = \frac{\sum_{i \in S_v \cap S_t} z_{v,i}\, z_{t,i}}{\|\mathbf{z}_v\|_2\,\|\mathbf{z}_t\|_2},
\label{eq:cos_support}
\end{equation}
with $S_m = \{i : z_{m,i} > 0\}$ for $m \in \{v, t\}$. The numerator is positive only on indices active in \emph{both} modalities, so jointly maximizing the cosine and minimizing the $L_1$ sparsity term biases each encoder to select a small, overlapping active support $S := S_v \cap S_t$ for paired inputs. Conditioned on this shared support, the reconstructions decompose as
\begin{equation}
\hat{\mathbf{A}}_v \approx \sum_{i \in S} z_{v,i}\, \mathbf{w}^{(v)}_i,\quad \hat{\mathbf{A}}_t \approx \sum_{i \in S} z_{t,i}\, \mathbf{w}^{(t)}_i,
\label{eq:decoder_pair}
\end{equation}
where $\mathbf{w}^{(m)}_i$ is the $i$-th column of $\mathbf{D}_m$. The pair $(\mathbf{w}^{(v)}_i,\,\mathbf{w}^{(t)}_i)$ is intended to encode the same semantic factor $i$ in two modality-specific subspaces, with $\mathbf{w}^{(t)}_i$ acting as a text-side residual-stream direction associated with concept $i$.

\paragraph{Expected operating regime.}
The LM head decodes the layer-$\ell$ residual stream via a (nearly) linear unembedding $W_U$, and $\mathbf{w}^{(t)}_i$ is learned by reconstruction against text-side hidden states, so it should lie approximately in a subspace the unembedding reads from. We therefore expect that injecting $\alpha\,\mathbf{w}^{(t)}_i$ into image-token states $\mathbf{H}^{\text{vis}}$ will bias the downstream generation toward concept $i$ at depths where $\mathbf{H}^{\text{vis}}$ has been routed into the language-aligned subspace by cross-attention---consistent with the layer-wise probing pattern (Fig.~\ref{fig:appendix_probe}) and with~\citet{venhoff2025visualrepresentationsmaplanguage}. Conversely, very deep layers are expected to be less amenable to this kind of additive injection, since $\mathbf{w}^{(t)}_i$ was not learned against pre-logit states; we revisit this when interpreting the Layer 30 results (Section~\ref{sec:results}).

\subsection{Direct Semantic Feature Discovery}

Well-trained SAEs exhibit \textit{feature splitting}~\citep{bricken2023monosemantic}, decomposing broad categories (e.g., \textit{food}) into fine-grained atomic features (e.g., \textit{vegetables}, \textit{plates}). For macro-category steering, we re-aggregate atomic features into overarching concepts (Algorithm~\ref{alg:feature_discovery}): we filter inactive neurons by a variance threshold, build an $L_2$-normalized activation profile $\mathbf{a}_i \in \mathbb{R}^N$ per neuron, and cluster these via K-Means. Clusters are ranked by mean Silhouette score; we then qualitatively inspect top-ranked clusters to discard polysemantic ones and keep the top-$K$ as final alignment features $\mathcal{F}_{\text{aligned}}$.

\begin{algorithm}[tb]
   \caption{Direct Semantic Feature Discovery via Activation Profiling}
   \label{alg:feature_discovery}
\begin{algorithmic}
   \STATE {\bfseries Input:} Validation dataset of size $N$, paired latent activations $\mathbf{Z}_v, \mathbf{Z}_t \in \mathbb{R}^{N \times d_{\text{latent}}}$.
   \STATE {\bfseries Output:} Sparse alignment features $\mathcal{F}_{\text{aligned}}$ (Top-$K$ semantic categories).
   \STATE 1. \textbf{Active Neuron Filtering:} Compute activation variance $\sigma^2_i = \text{Var}(\mathbf{z}_{*,i})$. Retain candidate indices $\mathcal{N} = \{i : \sigma^2_i > 10^{-5}\}$.
   \STATE 2. \textbf{Activation Profiling \& Normalization:} For each candidate $i \in \mathcal{N}$, extract its activation profile $\mathbf{a}_i \in \mathbb{R}^N$. Apply $L_2$-normalization: $\tilde{\mathbf{a}}_i = \mathbf{a}_i / \|\mathbf{a}_i\|_2$.
   \STATE 3. \textbf{Functional Clustering:} Perform K-Means on the normalized profile vectors $\{\tilde{\mathbf{a}}_i\}_{i \in \mathcal{N}}$ to partition neurons into semantic clusters $\{C_1, \dots, C_M\}$.
   \STATE 4. \textbf{Structural Evaluation:} Calculate the mean Silhouette score $s_k$ for each cluster $C_k$ to evaluate geometric compactness.
   \STATE 5. \textbf{Semantic Validation \& Selection:} Rank clusters by $s_k$. Inspect the top $N$-highest activating samples for the top-ranked cluster centroids. Retain monosemantic clusters and select the top-$K$ as the final alignment features: $\mathcal{F}_{\text{aligned}} = \{C_{(1)}, \dots, C_{(K)}\}$.
\end{algorithmic}
\end{algorithm}

\subsection{Causal Intervention Analysis}
\label{sec:causal}

We probe whether the discovered features influence cross-modal generation via additive activation steering. Because JSAE's alignment loss encourages directional agreement between $\mathbf{z}_v$ and $\mathbf{z}_t$, the text-decoder columns $\mathbf{W}_{\text{dec}}^{(t)}$ serve as approximate steering directions for the visual stream.

\paragraph{Steering Procedure.}
As in Fig.~\ref{fig:main_architecture} (Stage~2), to intervene on a category (e.g., \textit{food}), we retrieve its atomic features $\mathcal{I}$ and aggregate their normalized text-decoder directions into a macro-concept vector:
\begin{equation}
\mathbf{v}_{\text{steer}} = \sum_{i \in \mathcal{I}} \frac{\mathbf{w}_i^{(t)}}{\|\mathbf{w}_i^{(t)}\|_2 + \epsilon},\quad \mathbf{H}' = \mathbf{H} + \alpha \cdot \mathbf{v}_{\text{steer}} \odot \mathbf{m}
\end{equation}
where $\mathbf{m}$ restricts the intervention to image-token positions, and $\alpha$ is calibrated via grid search to balance semantic impact and fluency.

\paragraph{Semantic Suppression (Inhibition).}
We apply the steering vector in the negative direction ($\mathbf{H}' = \mathbf{H} - \alpha \cdot \mathbf{v}_{\text{steer}} \odot \mathbf{m}$): if subtracting a direction causes a visually present concept to vanish from the caption, we treat it as evidence that the direction contributes to maintaining that grounded concept.

\section{Experiments}

\subsection{Experimental Setup}
\label{setup}
\paragraph{Datasets.}
From MS-COCO~\citep{chen2015microsoftcococaptionsdata} we build two non-overlapping splits. \textit{COCO-Probe} (25k image-caption pairs from 10k images, 2:2:1 ratio): 10k positive pairs, 10k hard negatives (most-similar caption from a different image via $L_2$-normalized BERT~\citep{devlin2019bertpretrainingdeepbidirectional} \texttt{[CLS]} cosine similarity), and 5k easy random mismatches. \textit{JSAE Training} (50k positive-only image-caption pairs from the COCO training split). Probe uses the COCO validation split to prevent contamination.

\paragraph{Models and Hardware.}
We use LLaVA-v1.6-Mistral-7B, Qwen3-VL-30B-Instruct (MoE), and Llama3-LLaVA-Next-8B, on NVIDIA A100 (80GB).

\paragraph{Layer Selection via Probing.}
Linear probes are trained per layer (mean-pooled $d{=}4096$ hidden states) for 10 epochs with cross-entropy and Adam (lr~$10^{-3}$) on an 80/20 split.

\paragraph{JSAE Training Protocol.}
JSAEs use $d{=}4096$, $k{=}16384$ ($4\times$ overcomplete), trained for 20 epochs with Adam (lr~$10^{-5}$). Loss coefficients: $\lambda_r{=}1.0$, $\lambda_s{=}0.03$, $\lambda_a{=}1.0$ (validation-tuned). Convergence: MSE $<5\times10^{-4}$, L0 sparsity $\lesssim 4\%$ ($\sim$100--700 active features out of 16{,}384), alignment loss $<10^{-4}$. Each SAE takes $\sim$6 A100-hours; the total budget across all SAE runs and interventions is $\sim$100 A100-hours.

\paragraph{Intervention Setup and Model Generalization.}
Our primary layer-wise analysis on LLaVA-v1.6-Mistral-7B spans Layers 7, 13, 25, and 30 (early/mid/late/pre-output); architectural generalization is tested on Qwen3-VL-30B-Instruct (MoE, Layer 33) and Llama3-LLaVA-Next-8B (Layer 15). Target categories are selected per layer based on observed semantic specialization, with concept-feature pairs and per-setting sample counts (28--40 held-out COCO images) listed in Table~\ref{tab:appendix-detailed-results}.

\paragraph{Baselines and Comparative Analysis.} To evaluate the intervention effects of our discovered features and the contribution of the joint alignment objective, we compare JSAE against four baselines and one ablation:

\begin{itemize}
    \item \textbf{Prompt Steering:} A behavioral baseline using prepended textual instructions (e.g., ``Describe the food'') to measure the model's inherent compliance without internal intervention.
    \item \textbf{Random Noise:} Injects Gaussian vectors with norms matched to our steering vectors. This verifies that the observed effects stem from specific semantic directions rather than general perturbation artifacts.
    \item \textbf{Activation Addition (ActAdd):} An optimization-free method~\citep{turner2024steeringlanguagemodelsactivation} that computes a steering vector via the algebraic mean difference between positive and negative example activations, serving as a heuristic baseline.
    \item \textbf{VL-SAE (Standard VLM SAE):} A standard SAE trained directly on the VLM's residual stream~\citep{shen2025vlsaeinterpretingenhancingvisionlanguage}. Unlike JSAE, it treats activations as a single distribution and lacks explicit cross-modal alignment constraints. For a fair comparison, we train VL-SAE on the same Layer 13 activations with matched hyperparameters (dictionary size $k{=}16384$, $\lambda_r{=}1.0$, $\lambda_s{=}0.03$, Adam, identical schedule); it converges to val reconstruction MSE $=8.0\times10^{-4}$ and $L_0 \approx 1.56\%$ ($\sim$256/16384 active), comparable to JSAE's reconstruction-sparsity regime, ruling out under-fitting as the cause of its zero-success steering.
    \item \textbf{JSAE w/o Alignment ($\lambda_a=0$):} An ablation study where the alignment loss is removed. This tests whether the learned sparse features alone, without forced cross-modal binding, are sufficient for effective steering.
\end{itemize}

\subsection{Evaluation Metrics and Protocol}

\paragraph{Evaluation Metrics.}
We use three complementary metrics. \textbf{Similarity Gain ($\Delta_{\text{sim}}$)} is the cosine-similarity shift toward the target concept in SBERT (\emph{all-MiniLM-L6-v2}) space:
\begin{equation}
    \Delta_{\text{sim}} = \cos(\mathbf{e}_{\text{steer}}, \mathbf{e}_{\text{target}}) - \cos(\mathbf{e}_{\text{base}}, \mathbf{e}_{\text{target}})
\end{equation}
with $\mathbf{e}_{\text{target}}$ encoding a target description (e.g., ``\textit{A photo of a variety of food}''). \textbf{Accuracy} (target-concept presence) and \textbf{Coherence} (logical consistency, freedom from hallucination) are scored on $1$--$10$ by \texttt{claude-sonnet-4} (prompts in Appendix~\ref{sec:appendix_prompts}). \textbf{Fluency} is the PPL ratio $r_{\text{PPL}}{=}\text{PPL}_{\text{steer}}/\text{PPL}_{\text{base}}$ (values near $1.0$ preserve fluency). These three metrics are produced by independent scoring functions (SBERT cosine, an LLM judge, and base-model perplexity), so $\Delta_{\text{sim}}$ and $r_{\text{PPL}}$ act as automated, judge-free corroboration for the Claude-scored Accuracy/Coherence.

\paragraph{Success Criteria and Statistical Testing.}
We define an individual intervention as successful if $\Delta_{\text{sim}}>0.2$, i.e.\ the steered caption's SBERT similarity to the target description rises by at least $0.2$ over the base caption. The Success Ratio is the percentage of test samples meeting this threshold; it is therefore an automated, judge-free statistic. As an empirical check, it aligns strongly with the Claude-scored Accuracy: on the LLaVA-7B cross-category steering set ($n{=}420$ pooled over Layers $\{7,13,25,30\}$), Spearman $\rho{=}{+}0.68$ and Pearson $r{=}{+}0.63$ (both $p{<}10^{-60}$), with $76.4\%$ per-sample agreement between $\Delta_{\text{sim}}{>}0.2$ and Accuracy${>}5$ ($88.5\%$ at Layer~13, where neither metric saturates; full per-layer statistics in Appendix~Table~\ref{tab:appendix-corr-simgain-accuracy}). Layer- and method-level significance is further assessed by Welch's $t$-tests on the underlying $\Delta_{\text{sim}}$ distribution (Section~\ref{sec:results}, Appendix~Table~\ref{tab:appendix-layerwise-ttest}).

\section{Results}
\label{sec:results}
Before assessing interventions, we verify that JSAE preserves the vision-side captioning signal under a stringent \emph{direct-replacement} protocol (all vision tokens overwritten by the broadcast JSAE reconstruction $H^{\text{vis}} = \mathbf{D}_v(\mathbf{E}_v(\overline{H}_{\text{vis}}))$). On $1{,}000$ COCO samples (content-word CE), recovery rises monotonically from $\sim$37\% at Layer~7 to $\sim$100\% by Layer~30, with JSAE matching or surpassing a raw-mean broadcast control at every layer (Appendix~\ref{appendix:ce_recovered}). Vision-position content thus becomes largely dispensable by Layer~25, consistent with the asymmetric dynamics below.

\subsection{Layer-wise Evolution of Steering Effectiveness}
Table~\ref{tab:main_results} reports layer-wise intervention effects in LLaVA-v1.6-Mistral-7B; per-category breakdowns appear in Table~\ref{tab:appendix-detailed-results} and qualitative examples in Appendix~\ref{appendix:appendix_cases}. Dataset-level generalization is further validated on $1{,}500$ Flickr30k~\citep{young2014image} images spanning three scene categories ($500$ each), where JSAE Layer~13 steering remains effective with category-specific transfer degradation (Appendix Table~\ref{tab:appendix-flickr-random}).
\paragraph{Steering Effectiveness Rises with Depth.} 
Success Ratio rises from $0.400$ at Layer 7 to $0.740$ at Layer 13, peaking at Layer 25 ($\mathbf{0.798}$, Accuracy $7.22 \pm 2.57$). Welch's $t$-tests on $\Delta_{\text{sim}}$ over the cross-category set confirm this peak: Layer 25 significantly exceeds Layers 7, 13, and 30 with Bonferroni-corrected $p_{\text{bonf}}\le 3.1\times 10^{-3}$ (Cohen's $d$ up to $+1.69$ for L25 vs L30; full pairwise statistics in Appendix~Table~\ref{tab:appendix-layerwise-ttest}). Importantly, this peak does not hinge on the LLM-as-judge: Success Ratio is computed directly from SBERT-based $\Delta_{\text{sim}}$, and the Claude-scored Accuracy independently induces the identical layer ranking (L25 $>$ L13 $>$ L7 $>$ L30; Spearman $\rho{=}1.0$), confirming agreement between the automated and judge-based signals. This is consistent with high-level semantic directions being more readily steerable in late-but-pre-output layers.

\paragraph{Diminishing Steerability at the Output Boundary.}
Steerability is not unbounded: at Layer~30 the success ratio drops to $0.152$ and similarity gain to $0.08$, though coherence remains high ($6.81$). A \emph{post-hoc} reading consistent with the design rationale (Section~\ref{sec:method_jsae}) is that near the unembedding, hidden states become increasingly tied to the linear LM head; since $\mathbf{w}^{(t)}_i$ was learned against intermediate-layer text activations rather than pre-logit states, its component orthogonal to that final subspace is attenuated by the softmax, limiting the effect of additive injection. Manifold-preserving or norm-controlled interventions are left to future work.
\begin{figure}[t]
    \centering
    \includegraphics[width=0.48\textwidth]{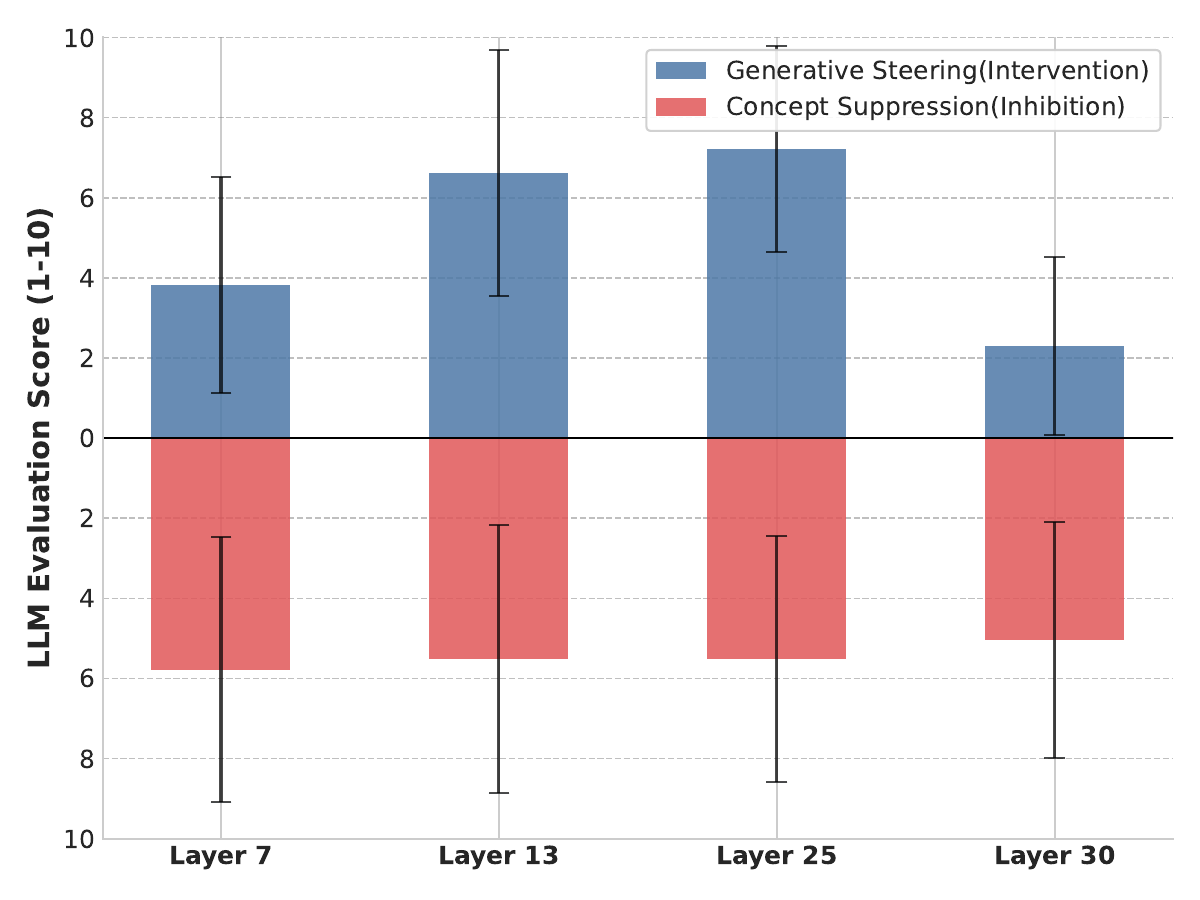}
    \caption{\textbf{Asymmetric Intervention Dynamics.} \textcolor{blue}{Upper (blue)}: additive steering Accuracy peaks at Layer 25 ($\approx 7.22$). \textcolor{red}{Lower (red)}: suppression scores stay in a narrow range ($\approx 5.5$) across depths. Additive controllability is layer-specific, while suppression effects appear across all probed layers.}
    \label{fig:bidirectional_causal}
\end{figure}

\paragraph{Stability of Suppression Effects.} 
Unlike the peaked trend of additive steering (Fig.~\ref{fig:bidirectional_causal}), suppression scores stay within a narrow range (Layer 7: $5.78 \pm 3.31$ to Layer 30: $5.04 \pm 2.94$), and all six pairwise Welch tests on suppression $\Delta_{\text{sim}}$ remain non-significant after Bonferroni correction (min $p_{\text{bonf}}{=}0.30$, max $|d|{=}0.39$; Appendix~Table~\ref{tab:appendix-layerwise-ttest}). This points to an asymmetry under our protocol: additive controllability concentrates in late-but-pre-output layers, while the identified directions appear behaviorally relevant across all probed depths.

\subsection{Baseline Comparisons and Ablation Studies}
We evaluate JSAE against baselines at the probe-optimal Layer 13 (Table~\ref{tab:main_results}, Part 4).

\paragraph{Role of Joint Alignment.} 
Both the unaligned ablation ($\lambda_a{=}0$) and VL-SAE perform poorly under our criterion ($\Delta_{\text{sim}}>0.2$): $0\%$ success for VL-SAE and $4.1\%$ for unaligned JSAE, with per-category Accuracy $\le 1.48$ (Table~\ref{tab:appendix-detailed-results}). This is not a training artefact---VL-SAE matches JSAE on reconstruction ($8.0\times10^{-4}$) and sparsity ($L_0{\approx}1.56\%$) under identical hyperparameters---but reflects a missing geometric property: without paired-code constraints, VL-SAE's shared dictionary columns reconstruct the mixed distribution rather than aligning directionally across modalities, so no column is bound to a cross-modal concept axis usable as a steering direction. The few hits in the unaligned case cluster in a single Surfing subset (Succ $0.129$, Acc $1.48{\pm}0.64$) whose Coherence ($4.78{\pm}3.03$) is the lowest---degenerate generations rather than concept transfer. The gap is direction- rather than scale-driven: rescaling unaligned $v_\text{steer}$ to $\|v_\text{aligned}\|$ leaves the statistics essentially unchanged (Acc $1.15$ vs.\ $1.14$; Succ $0.064$ vs.\ $0.041$), consistent with alignment contributing via cross-modal anchoring of dictionary directions rather than norm matching.
\paragraph{Comparison to Heuristics and Prompting.} 
JSAE attains $74.0\%$ success vs.\ $39.4\%$ for Activation Addition. A breakdown (Appendix Table~\ref{tab:appendix-detailed-results}) shows that ActAdd handles broad scenes well ($72.5\%$ on ``Indoor room'') but performs poorly on specific objects/actions ($12.5\%$ ``Wild animal''; $25.0\%$ ``Food''), suggesting that algebraic mean-difference vectors shift global context but less so fine-grained semantics. JSAE also substantially outperforms Prompt Steering ($22.1\%$) and Random Noise ($2.9\%$), indicating the observed effects are not driven by prompt-level instructions or generic perturbations.

\subsection{Generalization Across Architectures}
Given the per-layer training cost ($\sim$6 GPU-hours), we restrict each additional architecture to its probe-optimal layer (Table~\ref{tab:main_results}); the numbers should thus be read as evidence that \emph{at least one} steering-effective layer exists per architecture, not as the per-architecture optimum. JSAE transfers to the MoE-based \textbf{Qwen3-VL-30B} (Layer 33: $77.9\%$ Succ, Accuracy $7.02$, indicating steerable features under dynamic routing) and to \textbf{Llama3-LLaVA-8B} (Layer 15: $66.7\%$ Succ, Coherence $6.78$), suggesting that similar sparse alignment features arise in both dense and MoE VLMs tested here---though broader model coverage is needed for generality.

\begin{table*}[t]
\centering
\caption{Cross-architecture intervention results divided into four sections: (1) Layer-wise effects in LLaVA-v1.6-Mistral-7B; (2) Generalization on Qwen3-VL-30B; (3) Generalization on Llama3-LLaVA-8B; (4) Baseline comparisons and ablation studies on LLaVA-7B at Layer 13. The \textbf{Count} column indicates the number of valid test samples for each setting.}
\label{tab:main_results}
\resizebox{\textwidth}{!}{
\begin{tabular}{lcccccccc}
\toprule
\textbf{Model} & \textbf{Method} & \textbf{Layer} & \textbf{Count} & \textbf{Sim Gain} & \textbf{Accuracy} & \textbf{Coherence} & \textbf{Fluency (PPL)} & \textbf{Succ ratio} \\
\midrule
\multicolumn{9}{c}{\textit{\textbf{Part 1: LLaVA-7B Layer-wise Evolution}}} \\
\midrule
\textbf{LLaVA-7B} & \textbf{JSAE (Ours)} 
  & 7
    & 100 & $0.13 \pm 0.15$ & $3.82 \pm 2.70$ & $5.12 \pm 2.63$ & $0.91 \pm 0.52$ & $0.400$ \\
  \cmidrule{3-9}
  & & 13$^{\dagger}$
    & 104 & $0.34 \pm 0.22$ & $6.61 \pm 3.07$ & $\mathbf{6.94 \pm 1.77}$ & $1.06 \pm 0.36$ & $0.740$ \\
  \cmidrule{3-9}
  & & 25
    & 104 & $\mathbf{0.35 \pm 0.17}$ & $\mathbf{7.22 \pm 2.57}$ & $6.18 \pm 2.35$ & $1.10 \pm 0.35$ & $0.798$ \\
  \cmidrule{3-9}
  & & 30
    & 112 & $0.08 \pm 0.13$ & $2.29 \pm 2.22$ & $6.81 \pm 2.03$ & $1.09 \pm 0.29$ & $0.152$ \\
\midrule
\multicolumn{9}{c}{\textit{\textbf{Part 2: Qwen3-30B Generalization}}} \\
\midrule
\textbf{Qwen3-30B} & \textbf{JSAE} & 33$^{\dagger}$ 
    & 104 & $0.24 \pm 0.14$ & $7.02 \pm 2.74$ & $6.07 \pm 2.40$ & $1.00 \pm 0.30$ & $0.779$ \\
\midrule
\multicolumn{9}{c}{\textit{\textbf{Part 3: Llama3-8B Generalization}}} \\
\midrule
\textbf{Llama3-8B} & \textbf{JSAE} & 15$^{\dagger}$ 
    & 96 & $0.23 \pm 0.19$ & $6.14 \pm 3.49$ & $6.78 \pm 1.81$ & $1.08 \pm 0.45$ & $0.667$ \\
\midrule
\multicolumn{9}{c}{\textit{\textbf{Part 4: Baselines \& Ablation (on LLaVA-7B at Layer 13)}}} \\
\midrule
\textbf{LLaVA-7B} & Prompt & 13$^{\dagger}$
    & 104 & $0.04 \pm 0.08$ & $2.64 \pm 2.55$ & $7.49 \pm 1.39$ & $0.92 \pm 0.28$ & $0.221$ \\
  \cmidrule{2-9}
  & Random & 13$^{\dagger}$
    & 104 & $0.01 \pm 0.05$ & $1.42 \pm 1.19$ & $7.61 \pm 1.36$ & $0.98 \pm 0.25$ & $0.029$ \\
  \cmidrule{2-9}
  & VL-SAE & 13$^{\dagger}$
    & 112 & $0.00 \pm 0.05$ & $1.20 \pm 0.44$ & $7.42 \pm 1.25$ & $0.85 \pm 0.32$ & $0.000$ \\
  \cmidrule{2-9}
  & ActAdd & 13$^{\dagger}$
    & 104 & $0.26 \pm 0.12$ & $4.44 \pm 2.66$ & $7.55 \pm 1.40$ & $0.81 \pm 0.25$ & $0.394$ \\
  \cmidrule{2-9}
  & JSAE ($\lambda_a{=}0$) & 13$^{\dagger}$
  & 98 & $0.06 \pm 0.06$ & $1.14 \pm 0.40$ & $6.17 \pm 2.66$ & $0.93 \pm 0.30$ & $0.041$ \\
\bottomrule
\multicolumn{9}{l}{\footnotesize $^{\dagger}$ Indicates the probe-optimal layer (highest alignment score) selected for that specific architecture.} \\
\multicolumn{9}{l}{\footnotesize \textbf{Sim Gain}: Mean cosine similarity gain. \textbf{Accuracy/Coherence}: Claude-4-Sonnet scores (Mean $\pm$ Std). \textbf{Fluency}: PPL ratio.} \\
\multicolumn{9}{l}{\footnotesize \textbf{Succ ratio}: The percentage of successful interventions, where success is defined as $\Delta_{\text{sim}}>0.2$.} \\
\end{tabular}
}
\end{table*}

\section{Conclusion}
\label{conclusion}
We investigated vision-language alignment using Joint Sparse Autoencoders (JSAE), a framework that factorizes mean-pooled, image/caption-level cross-modal activations into sparse, intervention-amenable units. On LLaVA-7B, probing accuracy and steering controllability decouple: Layer 13 is probe-optimal, while Layer 25 yields the highest steering success ratio ($79.8\%$)---indicating that probe accuracy is not a sufficient predictor of steering effectiveness.

The two intervention modes show distinct depth profiles. Additive steering peaks at Layer 25 and drops near the unembedding layer, while suppression scores remain in a narrow range ($5.04$--$5.78$ out of 10) across all probed depths. The absolute suppression magnitude should not be over-interpreted in the absence of an unintervened or random-direction reference, but the lack of depth-localized collapse---including where additive steering weakens---is consistent with these directions remaining behaviorally relevant for concept grounding throughout the probed depths.

Two practical observations follow. First, the joint alignment objective matters: removing it ($\lambda_a{=}0$, $4.1\%$ Succ) or replacing JSAE with a standard VL-SAE ($0\%$) both fail to produce coherent steering. Second, learned sparse features outperform algebraic heuristics: JSAE substantially exceeds Activation Addition on fine-grained concepts and is more effective than Prompt Steering. Results on one dense and one MoE backbone suggest the pattern may extend beyond LLaVA, though broader model coverage is needed.

\section*{Limitations}
While JSAE provides evidence for informative hierarchical dynamics in vision-language alignment, several limitations remain (items marked ``future work'' are open problems, not claims).

\textbf{Granularity and Spatial Control:} JSAE relies on global mean-pooling to extract macroscopic representations, so our steering uniformly broadcasts semantic vectors across \textit{all} image tokens---acting as a global semantic filter rather than a localized object editor. Future work will explore token-centric alignment and attention-guided interventions for finer spatial control.

\textbf{Feature Aggregation vs. Monosemanticity:} Since SAEs naturally exhibit feature splitting, we employ functional clustering to aggregate atomic features into macro-concepts for category-level steering. An ideal representation space might instead map concepts hierarchically without post-hoc clustering; scaling dictionary sizes or adopting variants such as TopK SAEs may further refine dictionary purity.

\textbf{Scope and Applications:} Our evaluation focuses on single-layer interventions and high-level MS-COCO categories, with generalization further verified on $1{,}500$ Flickr30k~\citep{young2014image} images across three scenes (Appendix Table~\ref{tab:appendix-flickr-random}); the observed category-specific degradation (e.g., \textit{Indoor room}: $0.675 \to 0.340$) points to remaining domain sensitivity. Cross-layer dynamics, complex geometric reasoning, and cross-domain feature stability are left for future investigation, as is extending JSAE beyond generative steering to detect and inhibit multimodal hallucinations and unsafe concepts.
  
\section*{Impact Statement}
This paper presents a mechanistic interpretability framework for Vision-Language Models. While our primary goal is to better understand internal alignment, the steering capabilities studied here carry societal implications. On the positive side, the method could in principle support safety research by helping detect or inhibit harmful concepts or hallucinations once such concepts are reliably localized. On the negative side, like other steering techniques, it may also be misused to generate biased content. We view the analysis of these internal mechanisms as one component of broader efforts toward safer systems, and we encourage the community to explore safety-oriented circuit interventions.

\section*{Use of Scientific Artifacts}
All datasets and checkpoints are publicly released for research use under their respective licenses: MS-COCO (CC~BY~4.0; underlying Flickr image terms), Flickr30k (non-commercial research), and the LLaVA / Llama3-LLaVA / Qwen3-VL checkpoints and SBERT/BERT/NLTK tooling under their community or Apache-style licenses. Our use---probing, feature analysis, and activation steering---is consistent with their intended research use, and our released JSAE code follows the same. No new human data are collected; MS-COCO/Flickr30k images may incidentally depict individuals, but we perform no person identification or re-identification, and we did not observe offensive content in inspected samples.

\section*{Use of AI Assistants}
General-purpose AI assistants (GPT/Claude) were used for English polishing and minor code/debugging help. Independently, Claude Sonnet 4 serves as an automatic judge for Accuracy/Coherence (Section~\ref{setup}, Appendix~\ref{sec:appendix_prompts})---a methodological component, not a writing aid. All technical claims and experiments were verified by the authors.


\bibliography{acl}

\newpage
\appendix
\onecolumn

\section{LLM Evaluation Prompts}
\label{sec:appendix_prompts}

To ensure a robust evaluation of semantic alignment beyond surface-level N-gram metrics, we employed \texttt{claude-sonnet-4-20250514} as an impartial judge. Below, we provide the exact system prompts used to assess the \textbf{Generative Steering} (Targeted Intervention) and \textbf{Concept Suppression} (Same-Category Inhibition) tasks.

\subsection{Prompt for Generative Steering Evaluation}
This prompt is utilized to evaluate the success of steering the model to generate a specific target concept (e.g., ``food'') on an unrelated image (e.g., a landscape).

\begin{quote}
\small
\textbf{System Instruction:} You are an expert evaluator in Multimodal Learning and Computer Vision. Your task is to assess the quality of image captions generated by an AI model after a semantic intervention.

\vspace{0.5em}
\textbf{User Prompt:} \\
Below are the details of a generative steering experiment:
\begin{itemize}
    \item Target Concept: \texttt{\{target\_concept\}}
    \item Original Caption (Base): \texttt{\{original\_caption\}}
    \item Intervened Caption (Steered): \texttt{\{intervened\_caption\}}
\end{itemize}

Please evaluate the "Intervened Caption" based on the following two criteria using a scale of 1 to 10:

\begin{enumerate}
    \item \textbf{Semantic Accuracy:} How successfully does the Intervened Caption incorporate the core elements of the Target Concept? (1: No mention; 10: Perfect re-interpretation of the scene).
    \item \textbf{Textual Coherence:} Is the Intervened Caption grammatically correct, logically consistent, and free of unnatural hallucinations? (1: Gibberish; 10: Human-level fluency).
\end{enumerate}

\textbf{Output Format:}
\begin{itemize}
    \item Accuracy Score: [Score]
    \item Coherence Score: [Score]
    \item Reasoning: [Provide a concise explanation for your scores, focusing on how the visual features were re-mapped.]
\end{itemize}
\end{quote}

\subsection{Prompt for Concept Suppression Evaluation}
This prompt is utilized to evaluate the efficacy of suppressing a specific concept (e.g., dampening ``food'' features) when the image contains that concept.

\begin{quote}
\small
\textbf{System Instruction:} You are an expert evaluator in Multimodal Learning and Computer Vision. Your task is to assess the quality of image captions generated by an AI model after a semantic suppression intervention.

\vspace{0.5em}
\textbf{User Prompt:} \\
Below are the details of a generative suppression experiment:
\begin{itemize}
    \item Target Concept to Suppress: \texttt{\{target\_concept\}}
    \item Original Caption (Base): \texttt{\{original\_caption\}}
    \item Intervened Caption (Suppressed): \texttt{\{intervened\_caption\}}
\end{itemize}

Please evaluate the "Intervened Caption" based on the following two criteria using a scale of 1 to 10:

\begin{enumerate}
    \item \textbf{Suppression Effectiveness:} How successfully does the Intervened Caption reduce or eliminate references to the Target Concept while maintaining scene coherence? (1: Target concept still dominates; 10: Target concept completely removed/minimized without loss of meaning).
    \item \textbf{Textual Coherence:} Is the Intervened Caption grammatically correct, logically consistent, and free of unnatural hallucinations? (1: Gibberish; 10: Human-level fluency).
\end{enumerate}

\textbf{Output Format:}
\begin{itemize}
    \item Suppression Score: [Score]
    \item Coherence Score: [Score]
    \item Reasoning: [Provide a concise explanation for your scores, focusing on how the target concept was weakened or replaced while preserving caption quality.]
\end{itemize}
\end{quote}

\section{Detailed Experimental Results and Visualization}
\label{appendix:details}

\begin{figure}[t]
\centering
\includegraphics[width=1.0\columnwidth]{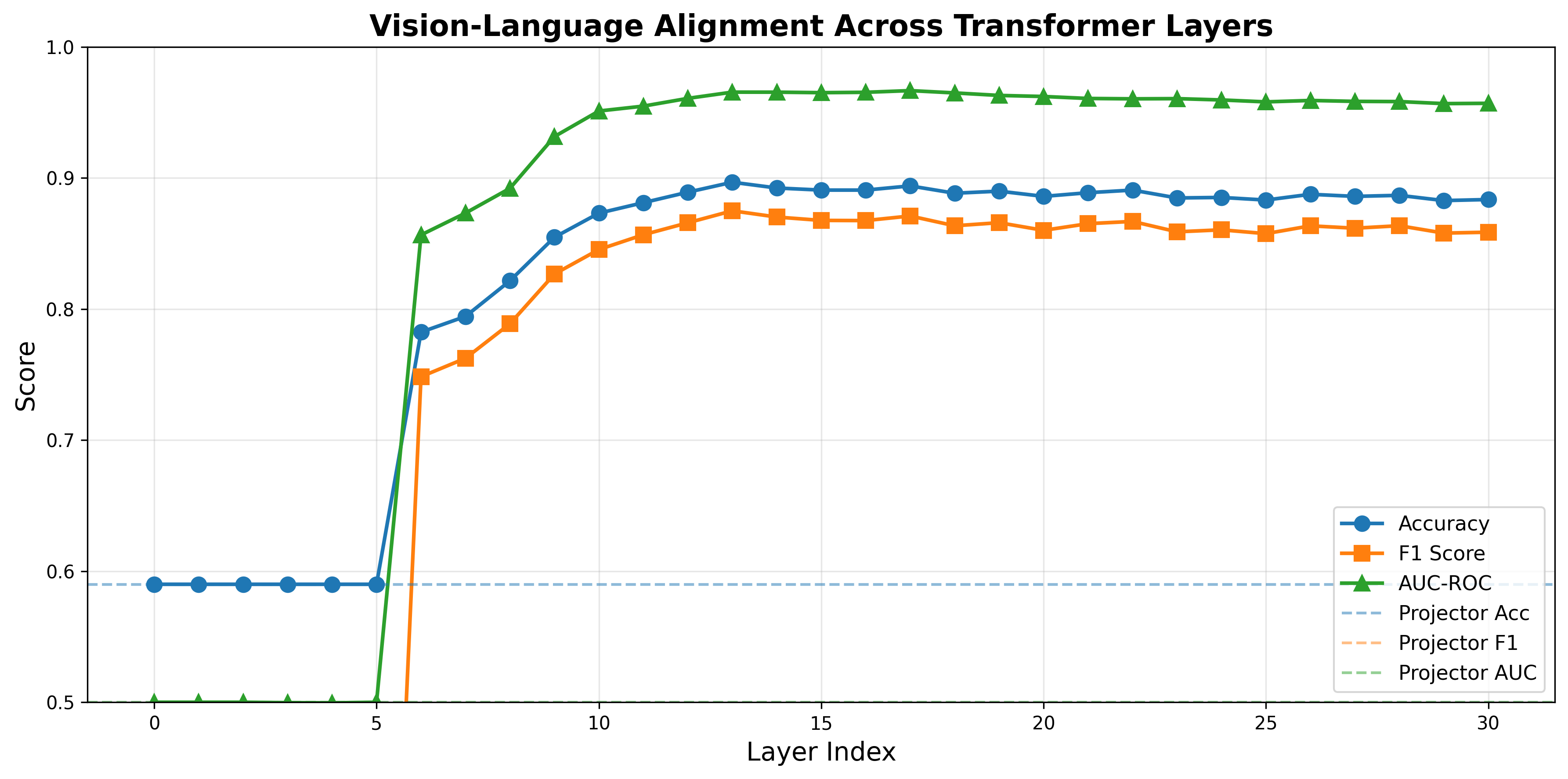}
\caption{F1 scores of layer-wise probe classifiers trained to detect text–image matching in LLaVA-v1.6-Mistral-7B using matched, simple mismatched, and hard mismatched samples, showing Layer~13 as the strongest multimodal alignment layer.}
\label{fig:appendix_probe}
\end{figure}

\begin{figure*}[ht]
    \centering
    \begin{subfigure}[b]{0.48\textwidth}
        \centering
        \includegraphics[width=\textwidth]{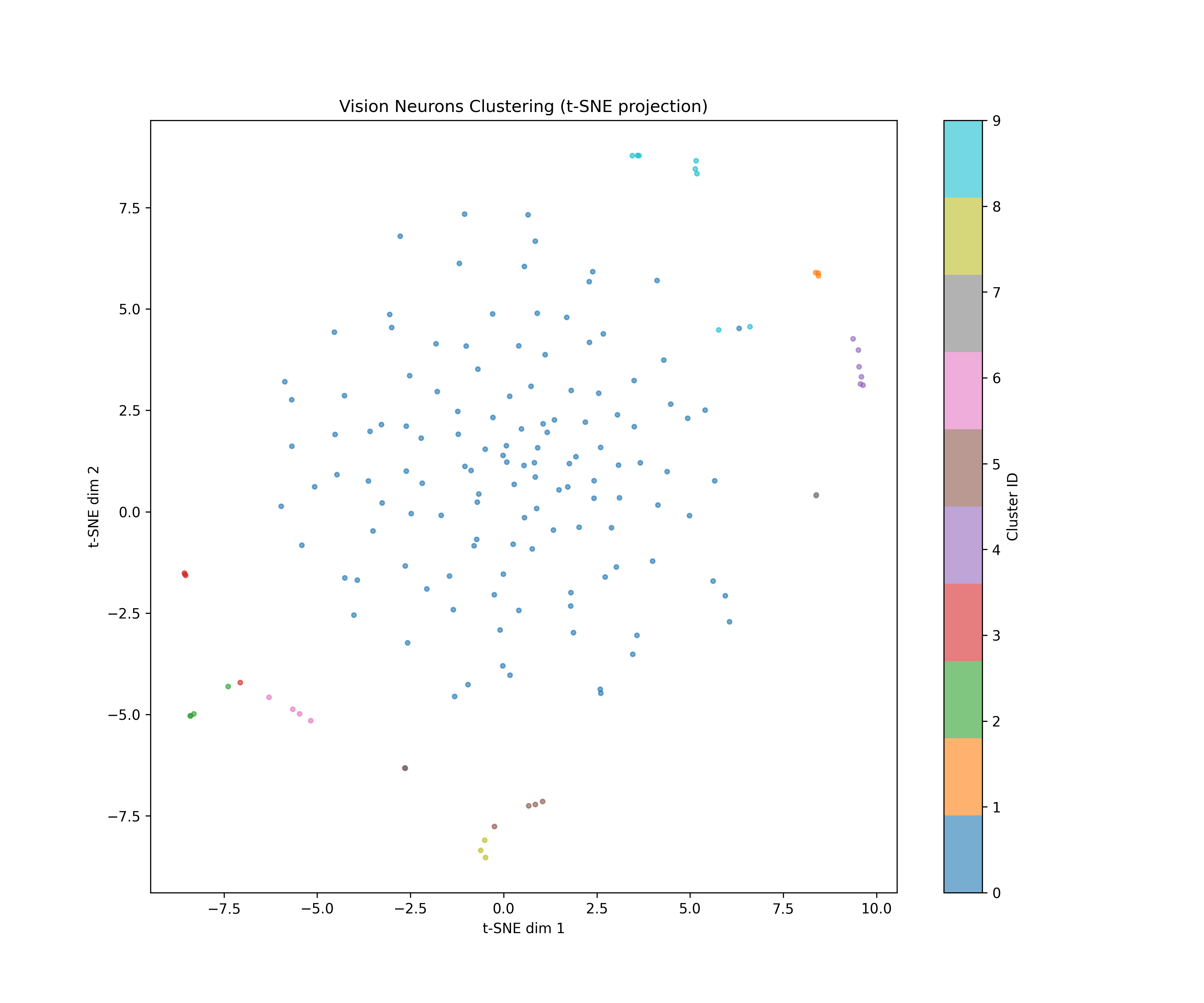}
        \caption{Layer 7}
        \label{fig:cluster_l7}
    \end{subfigure}
    \hfill
    \begin{subfigure}[b]{0.48\textwidth}
        \centering
        \includegraphics[width=\textwidth]{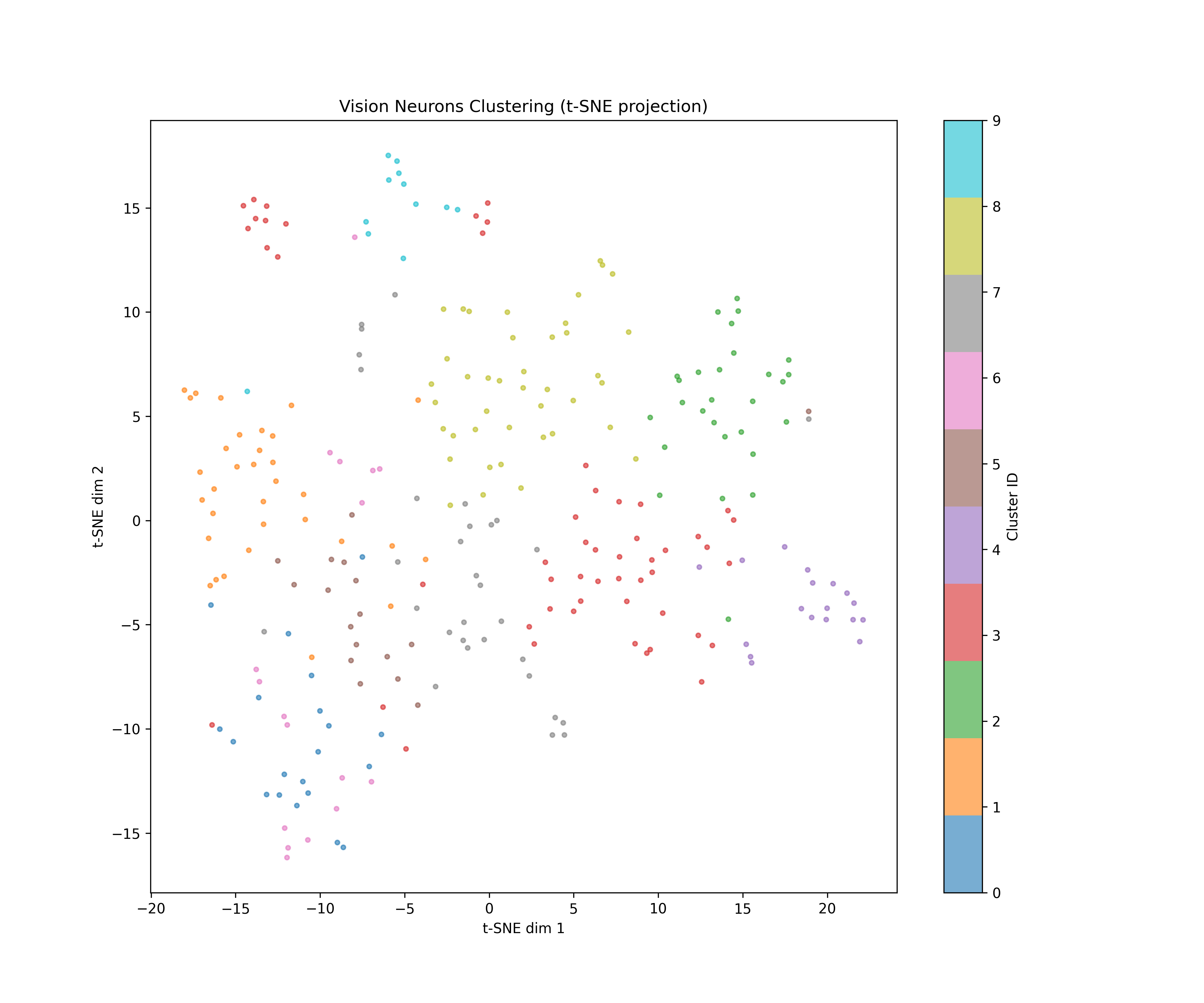}
        \caption{Layer 13}
        \label{fig:cluster_l13}
    \end{subfigure}
    
    \vspace{1em} 
    
    \begin{subfigure}[b]{0.48\textwidth}
        \centering
        \includegraphics[width=\textwidth]{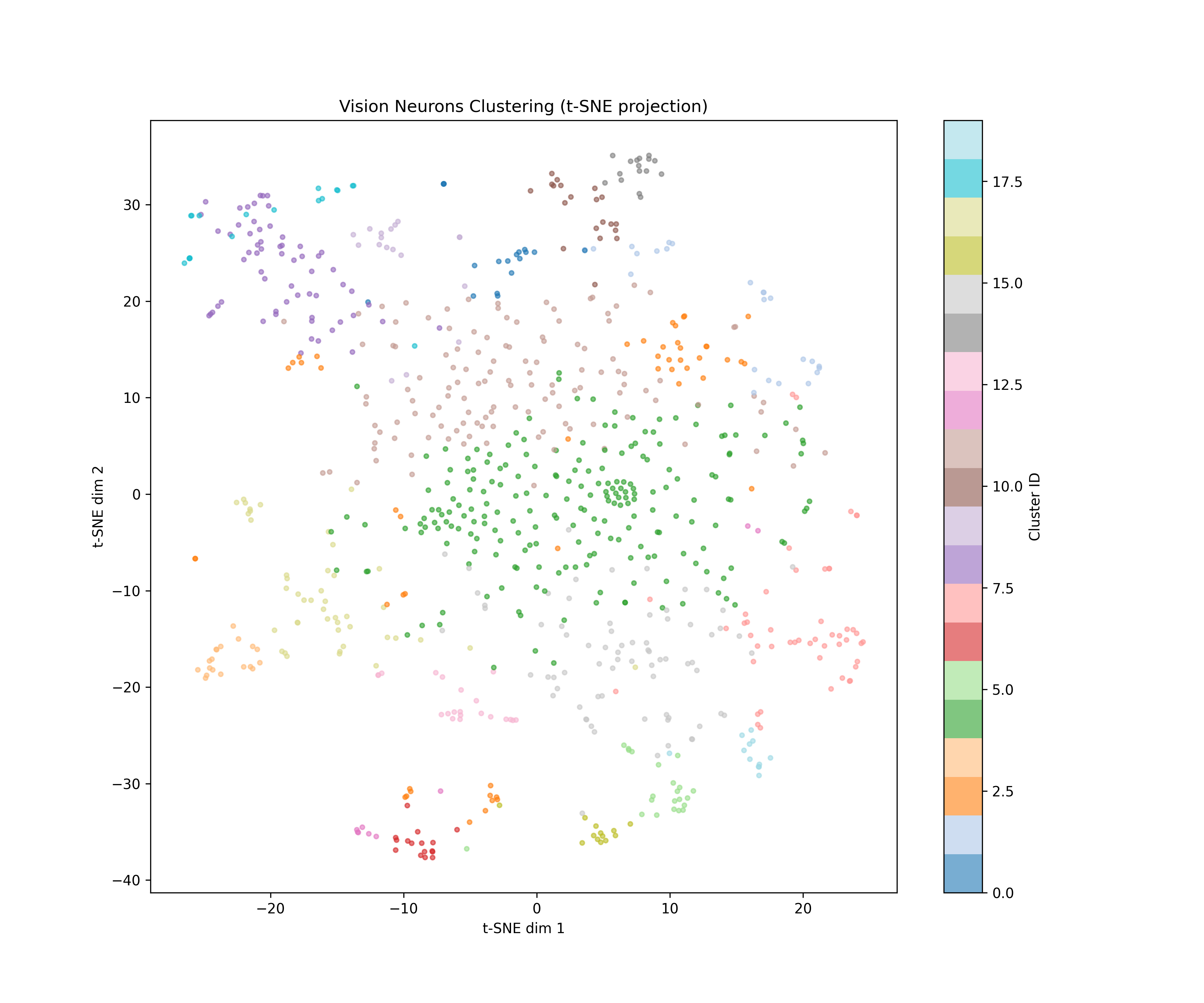}
        \caption{Layer 25}
        \label{fig:cluster_l25}
    \end{subfigure}
    \hfill
    \begin{subfigure}[b]{0.48\textwidth}
        \centering
        \includegraphics[width=\textwidth]{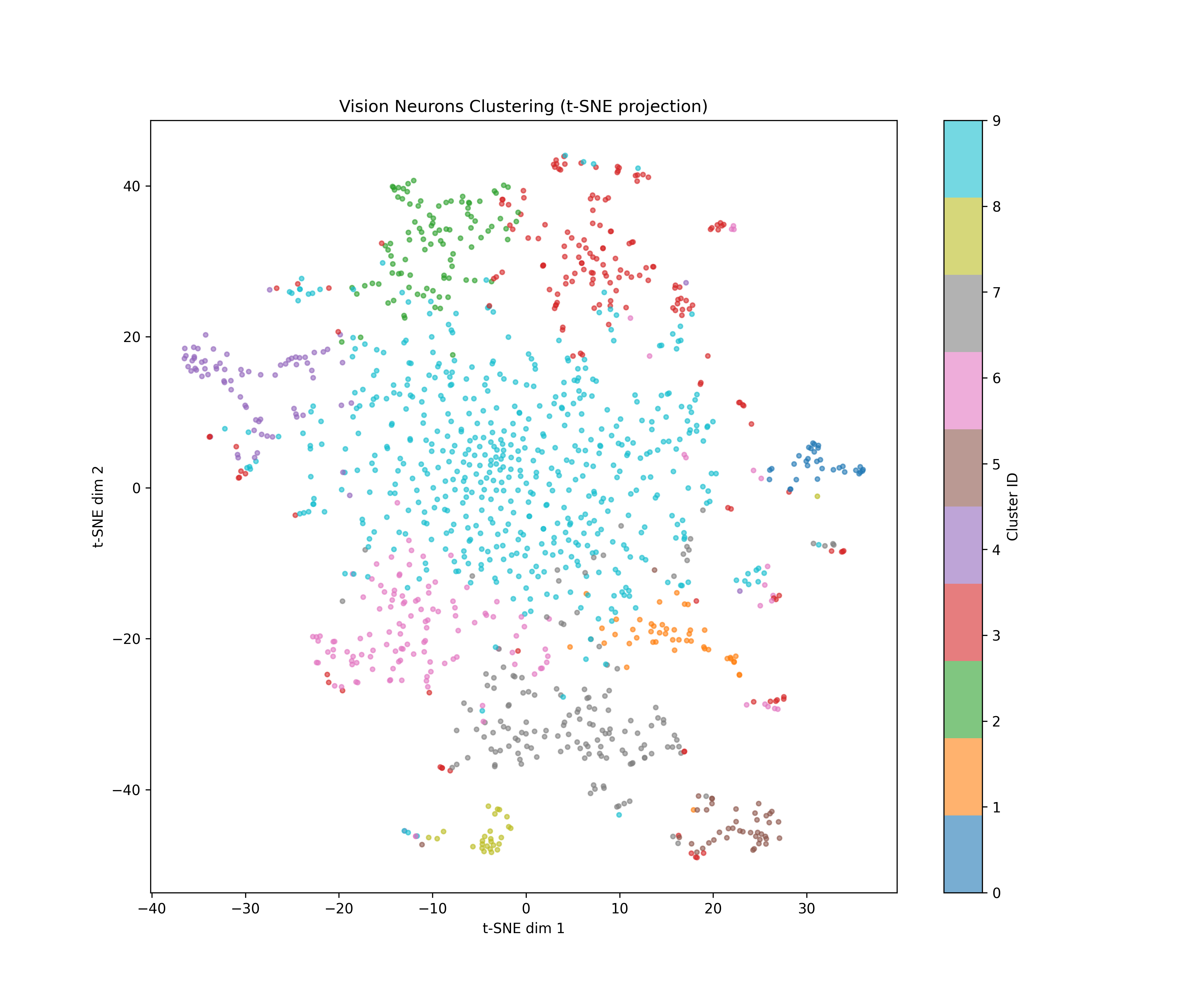}
        \caption{Layer 30}
        \label{fig:cluster_l30}
    \end{subfigure}
    
    \caption{t-SNE visualization of vision neuron clustering across Layers 7, 13, 25, and 30 in LLaVA-v1.6-7B. Each point represents a vision neuron, colored by its cluster ID. As depth increases from Layer 7 to Layer 25, neurons exhibit clearer semantic grouping, consistent with the increasing steering effectiveness observed in our intervention experiments. Layer 30 illustrates the representation state immediately preceding the output bottleneck.}
    \label{fig:appendix_clustering_all}
\end{figure*}

This section provides granular data and visual evidence for our hierarchical analysis. \textbf{Table~\ref{tab:appendix-detailed-results}} presents a full breakdown of intervention metrics across all models, layers, and categories. We first detail the linear probing analysis that guided our layer selection in \textbf{Figure~\ref{fig:appendix_probe}}, followed by the visualization of semantic structure in \textbf{Figure~\ref{fig:appendix_clustering_all}}.

\paragraph{Layer-wise Linear Probing.}
To empirically locate the layers where cross-modal alignment is most distinct, we trained linear classifiers (probes) on the residual streams of LLaVA-v1.6-7B to predict semantic concepts. As shown in \textbf{Figure~\ref{fig:appendix_probe}}, the probing accuracy rises sharply in the early layers and reaches a high plateau between Layer 13 and Layer 25. This trend implies that visual semantic information is maximally decodable and linearly separable in these middle-to-late layers. These findings justify our selection of Layers 13, 25, and 30 for JSAE training, as they capture the critical stages of the alignment process.

\paragraph{Hierarchical Trends and Failure Modes.} 
The data shows concept- and layer-dependent variation in intervention success rate. For example, LLaVA-7B's \textit{food} category success ratio escalates from $0.300$ (Layer 7) to $0.750$ (Layer 13). We also observe \emph{concept-specific failure modes}, notably for the \textbf{streets} category in Llama3-LLaVA-8B (Layer 15), which yielded a comparatively low success ratio of $0.237$ and minimal Similarity Gain ($0.05$) despite stable fluency. These observations are consistent with intervention success depending on both layer depth and concept type, but do not by themselves identify the underlying mechanism (e.g., dictionary purity vs.\ category-specific OOD effects).
\paragraph{Structural Evolution in Representation Space.}
The t-SNE projections in \textbf{Figure~\ref{fig:appendix_clustering_all}} visualize the semantic organization of neurons from Layer 7 through Layer 30. We observe that semantic clusters become increasingly distinct as depth increases, peaking around Layer 25. This structural pattern is consistent with the stronger additive steering effects reported in our main analysis. We also include Layer 30, showing that while structured clustering persists, the functional manipulability declines as the model approaches the output boundary.
\begin{table*}[t]
\centering
\caption{Detailed breakdown of intervention results across architectures and layers. We report the performance for the aggregated test set (Overall) followed by specific semantic categories to highlight concept-specific variations. \textbf{Count} denotes the number of test samples. All results correspond to cross-category (targeted) interventions.}
\label{tab:appendix-detailed-results}
\resizebox{\textwidth}{!}{
\begin{tabular}{llcccccccc}
\toprule
\textbf{Model} & \textbf{Category / Setting} & \textbf{Layer} & \textbf{Count} & \textbf{Sim Gain} & \textbf{Accuracy} & \textbf{Coherence} & \textbf{Fluency (PPL)} & \textbf{Succ ratio} \\
\midrule
\multicolumn{9}{c}{\textit{\textbf{Part 1: LLaVA-7B Layer-wise Evolution}}} \\
\midrule
\textbf{LLaVA-7B} & \textbf{Overall (JSAE)} & \textbf{7} & \textbf{100} & $\mathbf{0.13 \pm 0.15}$ & $\mathbf{3.82 \pm 2.70}$ & $\mathbf{5.12 \pm 2.63}$ & $\mathbf{0.91 \pm 0.52}$ & $\mathbf{0.400}$ \\
 & \hspace{1em} Food & & 30 & $0.12 \pm 0.16$ & $3.43 \pm 2.74$ & $4.73 \pm 2.41$ & $0.98 \pm 0.51$ & $0.300$ \\
 & \hspace{1em} Indoor room & & 38 & $0.11 \pm 0.13$ & $4.50 \pm 2.61$ & $4.13 \pm 2.42$ & $0.71 \pm 0.43$ & $0.553$ \\
 & \hspace{1em} Surfing & & 32 & $0.17 \pm 0.17$ & $3.38 \pm 2.70$ & $6.66 \pm 2.42$ & $1.10 \pm 0.56$ & $0.313$ \\
\cmidrule{2-9}
 & \textbf{Overall (JSAE)} & \textbf{13} & \textbf{104} & $\mathbf{0.34 \pm 0.22}$ & $\mathbf{6.61 \pm 3.07}$ & $\mathbf{6.94 \pm 1.77}$ & $\mathbf{1.06 \pm 0.36}$ & $\mathbf{0.740}$ \\
 & \hspace{1em} Food & & 32 & $0.31 \pm 0.22$ & $6.74 \pm 3.10$ & $6.74 \pm 1.98$ & $1.03 \pm 0.35$ & $0.750$ \\
 & \hspace{1em} Indoor room & & 40 & $0.26 \pm 0.21$ & $5.95 \pm 3.18$ & $6.95 \pm 1.56$ & $1.11 \pm 0.36$ & $0.675$ \\
 & \hspace{1em} Wild animal & & 32 & $0.47 \pm 0.19$ & $7.32 \pm 2.82$ & $7.13 \pm 1.84$ & $1.03 \pm 0.39$ & $0.813$ \\
\cmidrule{2-9}
 & \textbf{Overall (JSAE)} & \textbf{25} & \textbf{104} & $\mathbf{0.35 \pm 0.17}$ & $\mathbf{7.22 \pm 2.57}$ & $\mathbf{6.18 \pm 2.35}$ & $\mathbf{1.10 \pm 0.35}$ & $\mathbf{0.798}$ \\
 & \hspace{1em} Baseball & & 32 & $0.26 \pm 0.16$ & $7.22 \pm 2.96$ & $7.22 \pm 1.70$ & $1.13 \pm 0.36$ & $0.781$ \\
 & \hspace{1em} Skiing & & 34 & $0.42 \pm 0.17$ & $6.97 \pm 2.42$ & $5.00 \pm 2.49$ & $1.04 \pm 0.37$ & $0.794$ \\
 & \hspace{1em} Tennis & & 38 & $0.35 \pm 0.14$ & $7.45 \pm 2.38$ & $6.37 \pm 2.28$ & $1.13 \pm 0.32$ & $0.816$ \\
\cmidrule{2-9}
 & \textbf{Overall (JSAE)} & \textbf{30} & \textbf{112} & $\mathbf{0.08 \pm 0.13}$ & $\mathbf{2.29 \pm 2.22}$ & $\mathbf{6.81 \pm 2.03}$ & $\mathbf{1.09 \pm 0.29}$ & $\mathbf{0.152}$ \\
 & \hspace{1em} Baseball & & 36 & $0.24 \pm 0.13$ & $4.69 \pm 2.54$ & $5.25 \pm 2.42$ & $1.16 \pm 0.39$ & $0.472$ \\
 & \hspace{1em} Skiing & & 38 & $0.01 \pm 0.04$ & $1.11 \pm 0.31$ & $7.78 \pm 0.82$ & $1.07 \pm 0.23$ & $0.000$ \\
 & \hspace{1em} Surfing & & 38 & $0.01 \pm 0.04$ & $1.16 \pm 0.37$ & $7.34 \pm 1.58$ & $1.06 \pm 0.24$ & $0.000$ \\
\midrule
\multicolumn{9}{c}{\textit{\textbf{Part 2: Baselines on LLaVA-7B}}} \\
\midrule
\textbf{LLaVA-7B} & \textbf{VL-SAE (Overall)} & \textbf{13} & \textbf{112} & $\mathbf{0.00 \pm 0.05}$ & $\mathbf{1.20 \pm 0.44}$ & $\mathbf{7.42 \pm 1.25}$ & $\mathbf{0.85 \pm 0.32}$ & $\mathbf{0.000}$ \\
 & \hspace{1em} Food & & 36 & $-0.01 \pm 0.04$ & $1.17 \pm 0.38$ & $7.56 \pm 1.58$ & $0.88 \pm 0.41$ & $0.000$ \\
 & \hspace{1em} Skiing & & 38 & $0.01 \pm 0.06$ & $1.13 \pm 0.41$ & $7.24 \pm 1.00$ & $0.85 \pm 0.26$ & $0.000$ \\
 & \hspace{1em} Surfing & & 38 & $0.01 \pm 0.05$ & $1.29 \pm 0.52$ & $7.47 \pm 1.13$ & $0.81 \pm 0.26$ & $0.000$ \\
\cmidrule{2-9}
 & \textbf{ActAdd (Overall)} & \textbf{13} & \textbf{104} & $\mathbf{0.26 \pm 0.12}$ & $\mathbf{4.44 \pm 2.66}$ & $\mathbf{7.55 \pm 1.40}$ & $\mathbf{0.81 \pm 0.25}$ & $\mathbf{0.394}$ \\
 & \hspace{1em} Food & & 32 & $0.31 \pm 0.13$ & $3.81 \pm 2.48$ & $7.19 \pm 1.57$ & $0.88 \pm 0.22$ & $0.250$ \\
 & \hspace{1em} Indoor room & & 40 & $0.21 \pm 0.12$ & $6.23 \pm 2.45$ & $7.60 \pm 1.55$ & $0.78 \pm 0.22$ & $0.725$ \\
 & \hspace{1em} Wild animal & & 32 & $0.27 \pm 0.08$ & $2.84 \pm 1.63$ & $7.84 \pm 0.88$ & $0.78 \pm 0.30$ & $0.125$ \\
\cmidrule{2-9}
& \textbf{JSAE ($\lambda_a{=}0$)} & \textbf{13} & \textbf{98} & $\mathbf{0.06 \pm 0.06}$ & $\mathbf{1.14 \pm 0.40}$ & $\mathbf{6.17 \pm 2.66}$ & $\mathbf{0.93 \pm 0.30}$ & $\mathbf{0.041}$ \\
 & \hspace{1em} Airplanes & & 36 & $0.04 \pm 0.04$ & $1.00 \pm 0.00$ & $6.19 \pm 2.71$ & $1.06 \pm 0.30$ & $0.000$ \\
 & \hspace{1em} Skiing & & 31 & $0.04 \pm 0.03$ & $1.00 \pm 0.00$ & $7.35 \pm 1.50$ & $0.98 \pm 0.25$ & $0.000$ \\
 & \hspace{1em} Surfing & & 31 & $0.09 \pm 0.08$ & $1.48 \pm 0.64$ & $4.78 \pm 3.03$ & $0.74 \pm 0.26$ & $0.129$ \\
\midrule
\multicolumn{9}{c}{\textit{\textbf{Part 3: Cross-Architecture Generalization}}} \\
\midrule
\textbf{Qwen3-30B} & \textbf{Overall (JSAE)} & \textbf{33} & \textbf{104} & $\mathbf{0.24 \pm 0.14}$ & $\mathbf{7.02 \pm 2.74}$ & $\mathbf{6.07 \pm 2.40}$ & $\mathbf{1.00 \pm 0.30}$ & $\mathbf{0.779}$ \\
 & \hspace{1em} Food & & 32 & $0.33 \pm 0.13$ & $7.69 \pm 1.79$ & $5.22 \pm 2.43$ & $1.14 \pm 0.28$ & $0.906$ \\
 & \hspace{1em} Indoor room & & 40 & $0.18 \pm 0.09$ & $8.05 \pm 2.06$ & $7.13 \pm 2.02$ & $0.90 \pm 0.28$ & $0.875$ \\
 & \hspace{1em} Wild animal & & 32 & $0.20 \pm 0.16$ & $5.09 \pm 3.28$ & $5.63 \pm 2.39$ & $1.00 \pm 0.29$ & $0.531$ \\
\midrule
\textbf{Llama3-8B} & \textbf{Overall (JSAE)} & \textbf{15} & \textbf{96} & $\mathbf{0.23 \pm 0.19}$ & $\mathbf{6.14 \pm 3.49}$ & $\mathbf{6.78 \pm 1.81}$ & $\mathbf{1.08 \pm 0.45}$ & $\mathbf{0.667}$ \\
 & \hspace{1em} Food & & 28 & $0.29 \pm 0.06$ & $8.82 \pm 0.55$ & $6.21 \pm 1.64$ & $0.92 \pm 0.43$ & $1.000$ \\
 & \hspace{1em} Surfing & & 30 & $0.39 \pm 0.19$ & $7.73 \pm 2.24$ & $5.83 \pm 2.09$ & $1.35 \pm 0.53$ & $0.900$ \\
 & \hspace{1em} Streets & & 38 & $0.05 \pm 0.07$ & $2.89 \pm 2.96$ & $7.95 \pm 0.80$ & $0.98 \pm 0.26$ & $0.237$ \\
\bottomrule
\multicolumn{9}{l}{\footnotesize \textbf{Sim Gain}: Mean cosine similarity gain. \textbf{Accuracy/Coherence}: Claude-4-Sonnet scores. \textbf{Succ ratio}: $\Delta_{\text{sim}}>0.2$ rate.} \\
\end{tabular}
}
\end{table*}

\begin{table}[t]
    \centering
    \caption{Dataset-level generalization: Layer~13 JSAE steering on a Flickr30k~\citep{young2014image} subset of $1{,}500$ images ($500$ per category), unused for JSAE training or feature selection.}
    \label{tab:appendix-flickr-random}
    \resizebox{\columnwidth}{!}{
    \begin{tabular}{l c c c c c c c}
    \toprule
    \textbf{Category} & \textbf{Layer} & \textbf{Count} & \textbf{Sim Gain} & \textbf{Accuracy} & \textbf{Coherence} & \textbf{Fluency} & \textbf{Succ} \\
    \midrule
    Food         & 13 & 500 & $0.29 \pm 0.21$ & $4.80 \pm 2.94$ & $5.79 \pm 2.19$ & $1.29 \pm 0.55$ & $0.505$ \\
    Wild animal  & 13 & 500 & $0.37 \pm 0.23$ & $6.02 \pm 3.37$ & $7.11 \pm 1.54$ & $1.15 \pm 0.45$ & $0.620$ \\
    Indoor room  & 13 & 500 & $0.19 \pm 0.17$ & $3.69 \pm 2.92$ & $6.28 \pm 1.83$ & $1.38 \pm 0.60$ & $0.340$ \\
    \bottomrule
    \end{tabular}
    }
\end{table}

\section{Reconstruction Fidelity Details (CE Loss Recovered)}
\label{appendix:ce_recovered}

We provide the full per-layer breakdown of CE Loss Recovered in Table~\ref{tab:appendix-ce-recovered}. We adopt a stringent \emph{direct-replacement} variant of the standard protocol~\citep{templeton2024scaling}:
\begin{equation*}
\text{Recovered} = \frac{L_{\text{break}} - L_{\text{recon}}}{L_{\text{break}} - L_{\text{orig}}} \times 100\%,
\end{equation*}
where, at the target layer of an otherwise-clean forward pass, \emph{all} vision tokens are overwritten by the broadcast JSAE reconstruction $H^{\text{vis}} = \mathbf{D}_v(\mathbf{E}_v(\overline{H}_{\text{vis}}))$, with $\overline{H}_{\text{vis}}$ the mean-pooled vision activation. Unlike a residual-injection scheme that leaves per-patch detail intact, this protocol forces the model to caption purely from the SAE-compressed visual summary, providing a worst-case test of whether the autoencoder retains caption-relevant information. To isolate SAE reconstruction error from the inherent broadcast bottleneck, we additionally report a control $L_{\text{mean}}$ that overwrites vision tokens with the raw global mean (no SAE). CE losses are computed on noun/adjective tokens of each caption (parsed via NLTK \texttt{averaged\_perceptron\_tagger}) over $1{,}000$ randomly sampled COCO image--caption pairs.

\textit{Reading the table.} Sub-table (a) verifies that both corruption constructions are discriminative ($\Delta \ge +0.96$ nat). Sub-table (b) shows two findings. First, $L_{\text{recon}} \le L_{\text{mean}}$ at \emph{every} layer, with JSAE outperforming the raw-mean control by up to $13$\,pp.\ at early/mid layers and converging at late layers---indicating that the autoencoder introduces no reconstruction overhead and may even denoise the mean-pooled summary. Second, recovery rises monotonically with depth ($36.8\% \to 100.2\%$ under \texttt{global\_zero}; $47.7\% \to 100.2\%$ under \texttt{counterfactual}); by Layer~25, replacing all vision-position content with a single broadcast vector loses only $\le 5$\,pp.\ of caption signal, consistent with cross-modal attention having already aggregated visual information into the textual residual stream. This depth profile is consistent with the asymmetric steering dynamics observed in the main results.

\begin{table}[h]
\centering
\caption{CE Loss Recovered for JSAE reconstruction on LLaVA-v1.6-Mistral-7B ($1{,}000$ COCO samples, content-word CE). \textbf{(a)} Both corruption baselines increase CE ($\Delta \ge +0.96$ nat). \textbf{(b)} Under direct replacement, JSAE matches or surpasses a raw-mean broadcast control at every layer; recovery rises monotonically with depth, suggesting that vision-position content becomes less critical for the captioning signal at deeper layers under this protocol.}
\label{tab:appendix-ce-recovered}

\begin{tabular}{lcc}
\toprule
\multicolumn{3}{c}{\textbf{(a) Corruption baselines} (layer-agnostic, shared across all settings)} \\
\midrule
Setting & $L$ (nat) & $\Delta = L - L_{\text{orig}}$ \\
\midrule
$L_{\text{orig}}$ (clean inference) & $2.3348$ & --- \\
\texttt{global\_zero}                & $3.2951$ & $+0.9603$ \\
\texttt{counterfactual}              & $3.4955$ & $+1.1607$ \\
\bottomrule
\end{tabular}

\vspace{0.8em}

\begin{tabular}{lcc|cc|cc}
\toprule
\multicolumn{7}{c}{\textbf{(b) Per-layer reconstruction (direct replacement)}} \\
\midrule
 & \multicolumn{2}{c|}{Hidden state (nat)} & \multicolumn{2}{c|}{Recov$_{\texttt{gz}}$} & \multicolumn{2}{c}{Recov$_{\texttt{cf}}$} \\
Layer & $L_{\text{recon}}$ & $L_{\text{mean}}$ & JSAE & mean & JSAE & mean \\
\midrule
Layer 7  & $2.9422$ & $3.0639$ & $\phantom{0}36.8\%$ & $\phantom{0}24.1\%$ & $\phantom{0}47.7\%$ & $\phantom{0}37.2\%$ \\
Layer 13 & $2.8009$ & $2.8820$ & $\phantom{0}51.5\%$ & $\phantom{0}43.0\%$ & $\phantom{0}59.9\%$ & $\phantom{0}52.9\%$ \\
Layer 25 & $2.3800$ & $2.3849$ & $\phantom{0}95.3\%$ & $\phantom{0}94.8\%$ & $\phantom{0}96.1\%$ & $\phantom{0}95.7\%$ \\
Layer 30 & $2.3329$ & $2.3326$ & $100.2\%$           & $100.2\%$           & $100.2\%$           & $100.2\%$ \\
\bottomrule
\end{tabular}
\end{table}

\section{Layer-wise Statistical Significance}
\label{appendix:layerwise-ttest}

Table~\ref{tab:appendix-layerwise-ttest} reports complete pairwise Welch's $t$-tests on the SBERT similarity gain $\Delta_{\text{sim}}$ across all four probed layers of LLaVA-v1.6-Mistral-7B, for both intervention modes. Additive steering is computed on the cross-category subset (one steered run per image with the target concept differing from the image content; $n{=}100/104/104/112$ for layers $7/13/25/30$, matching the Count column of Table~\ref{tab:main_results}). Suppression is computed on the same-category subset (target concept matches the visually grounded one; $n{=}50/52/52/56$). Bonferroni correction uses $m{=}6$ comparisons per mode. The two modes display strongly contrasting patterns: every additive pair involving Layer~25 or Layer~30 survives correction at $p_{\text{bonf}}<6\times 10^{-3}$ with medium-to-very-large effects ($|d|\in[0.49,1.69]$), whereas \emph{no} suppression pair reaches significance (smallest $p_{\text{bonf}}{=}0.30$). This is the quantitative basis for the asymmetry summarized in Section~\ref{sec:results}.

\begin{table}[h]
\centering
\caption{Pairwise Welch's $t$-tests on $\Delta_{\text{sim}}$ across layers of LLaVA-v1.6-Mistral-7B. \textbf{Top}: additive steering (cross-category interventions). \textbf{Bottom}: same-category suppression. $p_{\text{bonf}}$ uses Bonferroni correction with $m{=}6$.}
\label{tab:appendix-layerwise-ttest}
\small
\begin{tabular}{lrrrrr}
\toprule
Pair & $\Delta\mu$ & $t$ & $p$ & $p_{\text{bonf}}$ & Cohen's $d$ \\
\midrule
\multicolumn{6}{c}{\textit{Additive steering (cross-category, $n_{7/13/25/30}{=}100/104/104/112$)}} \\
\midrule
L7 vs L13   & $-0.064$ & $-2.32$ & $2.1\times 10^{-2}$  & $1.3\times 10^{-1}$ & $-0.32$ \\
L7 vs L25   & $-0.159$ & $-6.56$ & $4.5\times 10^{-10}$ & $2.7\times 10^{-9}$ & $-0.92$ \\
L7 vs L30   & $+0.102$ & $+4.55$ & $9.6\times 10^{-6}$  & $5.7\times 10^{-5}$ & $+0.63$ \\
L13 vs L25  & $-0.094$ & $-3.53$ & $5.2\times 10^{-4}$  & $3.1\times 10^{-3}$ & $-0.49$ \\
L13 vs L30  & $+0.166$ & $+6.62$ & $4.2\times 10^{-10}$ & $2.5\times 10^{-9}$ & $+0.92$ \\
L25 vs L30  & $+0.261$ & $+12.32$ & $2.0\times 10^{-26}$ & $1.2\times 10^{-25}$ & $+1.69$ \\
\midrule
\multicolumn{6}{c}{\textit{Suppression (same-category, $n_{7/13/25/30}{=}50/52/52/56$)}} \\
\midrule
L7 vs L13   & $+0.007$ & $+0.21$ & $0.83$ & $1.00$ & $+0.04$ \\
L7 vs L25   & $+0.037$ & $+1.01$ & $0.31$ & $1.00$ & $+0.20$ \\
L7 vs L30   & $-0.017$ & $-0.50$ & $0.62$ & $1.00$ & $-0.10$ \\
L13 vs L25  & $+0.030$ & $+1.06$ & $0.29$ & $1.00$ & $+0.21$ \\
L13 vs L30  & $-0.024$ & $-0.99$ & $0.32$ & $1.00$ & $-0.19$ \\
L25 vs L30  & $-0.054$ & $-1.99$ & $4.96\times 10^{-2}$ & $0.30$ & $-0.39$ \\
\bottomrule
\end{tabular}
\end{table}

\section{Correlation Between Sim Gain and Judge Accuracy}
\label{appendix:corr-simgain-accuracy}

Table~\ref{tab:appendix-corr-simgain-accuracy} reports per-layer Pearson and Spearman correlations between $\Delta_{\text{sim}}$ and Claude-scored Accuracy on the LLaVA-v1.6-Mistral-7B cross-category steering set ($n{=}100/104/104/112$ for layers $7/13/25/30$, matching Table~\ref{tab:main_results}). The within-layer pattern reflects range compression at the extremes: alignment is strongest at Layer~13, where neither metric saturates, and is mechanically attenuated at Layer~25 (Sim Gain near a ceiling, Succ$=0.798$) and Layer~30 (judge Accuracy near a floor, Succ$(\text{Acc}{>}5){=}0\%$). The across-layer ranking is unaffected (Section~\ref{sec:results}).

\begin{table}[h]
\centering
\caption{Correlation between $\Delta_{\text{sim}}$ and Claude-scored Accuracy on LLaVA-v1.6-Mistral-7B cross-category steering interventions. Succ$(\Delta_{\text{sim}}{>}0.2)$ is taken from Table~\ref{tab:main_results} (Pooled row is the $n$-weighted average over the four layers). ``Agree'' is the rate at which the binary success criteria $\Delta_{\text{sim}}{>}0.2$ and Accuracy${>}5$ assign the same label, computed on the same judged outputs. $p$-values are two-sided.}
\label{tab:appendix-corr-simgain-accuracy}
\small
\begin{tabular}{lrrrrrrr}
\toprule
Set & $n$ & Pearson $r$ & $p_{\text{Pearson}}$ & Spearman $\rho$ & $p_{\text{Spearman}}$ & Succ$(\Delta_{\text{sim}}{>}0.2)$ & Agree \\
\midrule
Layer~7            & 100 & $+0.697$ & $6.8\times 10^{-22}$ & $+0.722$ & $5.6\times 10^{-25}$ & $0.400$ & $0.760$ \\
Layer~13           & 104 & $\mathbf{+0.847}$ & $\mathbf{4.0\times 10^{-58}}$ & $\mathbf{+0.888}$ & $\mathbf{4.3\times 10^{-85}}$ & $0.740$ & $\mathbf{0.885}$ \\
Layer~25           & 104 & $+0.180$ & $6.4\times 10^{-2}$ & $+0.152$ & $1.2\times 10^{-1}$ & $0.798$ & $0.644$ \\
Layer~30           & 112 & $+0.251$ & $6.7\times 10^{-3}$ & $+0.160$ & $8.9\times 10^{-2}$ & $0.152$ & $0.768$ \\
\midrule
Pooled (4 layers)  & 420 & $+0.627$ & $8.7\times 10^{-61}$ & $+0.677$ & $6.5\times 10^{-79}$ & $0.517$ & $0.764$ \\
\bottomrule
\end{tabular}
\end{table}

\section{Qualitative Analysis of Generative Steering}
\label{appendix:appendix_cases}

To further illustrate the steering effects of the identified circuits, we present representative cases of cross-category steering across different models and layers. These examples show how intervening on sparse neuron pairs can shift a model's generated description toward a target semantic category.

\subsection{Case Study 1: Tennis-to-Baseball (LLaVA-v1.6-7B, Layer 25)}
In this experiment using LLaVA-v1.6-7B at Layer 25, we amplify neurons aligned with the \textit{baseball} category while providing an image of a tennis player. As shown in Figure~\ref{fig:steering_tennis_baseball}, the generated caption describes the racket as a bat and the court as a field, while preserving some attributes of the original yellow-shirted subject.

\begin{figure*}[ht]
    \centering
    \begin{minipage}{0.3\textwidth}
        \centering
        \includegraphics[width=\textwidth]{exp1.png}
        \caption*{Input Image (Tennis)}
    \end{minipage}
    \hfill
    \begin{minipage}{0.65\textwidth}
        \small
        \textbf{Base Generation (Original):}\\
        \textit{"The image shows a tennis player on a blue hard court. The player is wearing a yellow shirt, black shorts, and white shoes. He is holding a tennis racket in his right hand and appears to be taking a break from play, as he..."}
        
        \vspace{1.5em}
        
        \textbf{Steered Generation (Target Category: Baseball):}\\
        \textit{"The image shows a scene from a baseball game. In the foreground, a baseball player is standing on the field, wearing a yellow jersey with blue sleeves, black pants, and white cleats. He is holding a baseball bat and appears..."}
    \end{minipage}
    \vspace{1em}
    \caption{Example of cross-category steering at Layer 25 of LLaVA-v1.6-7B. The generated caption shifts tennis-specific visual details toward the target baseball category while preserving base subject attributes like color.}
    \label{fig:steering_tennis_baseball}
\end{figure*}

\subsection{Case Study 2: Wild Animal to Food (LLaVA-v1.6-7B, Layer 13)}

\begin{figure*}[ht]
    \centering
    \begin{minipage}{0.3\textwidth}
        \centering
        \includegraphics[width=\textwidth]{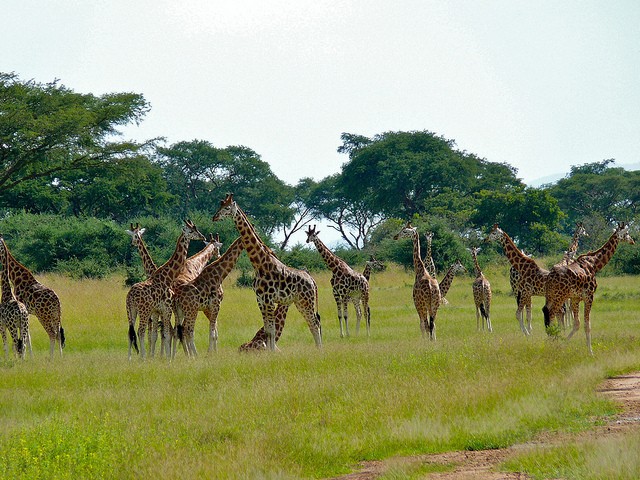}
        \caption*{Input Image (Giraffes)}
    \end{minipage}
    \hfill
    \begin{minipage}{0.65\textwidth}
        \small
        \textbf{Base Generation (Original):}\\
        \textit{"The image depicts a group of giraffes in a grassy field. There are at least nine giraffes visible, with varying shades of brown and white on their coats. They are standing and walking in different directions, some facing..."}
        
        \vspace{1.5em}
        
        \textbf{Steered Generation (Target Category: Food):}\\
        \textit{"The image shows a variety of savory dishes, including grilled chicken, vegetables, and a side of rice. The dishes are served on a bed of greens, and the meal is garnished with a sprinkle of herbs. The..."}
    \end{minipage}
    
    \vspace{1em}
    \caption{Example of cross-category steering at Layer 13 of LLaVA-v1.6-7B. The intervention re-directs the model to map the organic patterns and textures of the giraffes onto culinary concepts such as grilled proteins, while the green field is re-described as a bed of garnished greens.}
    \label{fig:steering_giraffe_food}
\end{figure*}

As shown in Figure~\ref{fig:steering_giraffe_food}, this case shows a biological/nature-oriented scene being described in a culinary context after an intermediate-depth intervention. By intervening on the \textit{food}-aligned sparse circuits, the generated caption maps visual patterns from giraffes and grass onto food-related descriptions such as savory dishes and garnishments.

\subsection{Case Study 3: Indoor Social Scene to Wild Animal (Qwen3-VL-30B, Layer 33)}

\begin{figure*}[ht]
    \centering
    \begin{minipage}{0.3\textwidth}
        \centering
        \includegraphics[width=\textwidth]{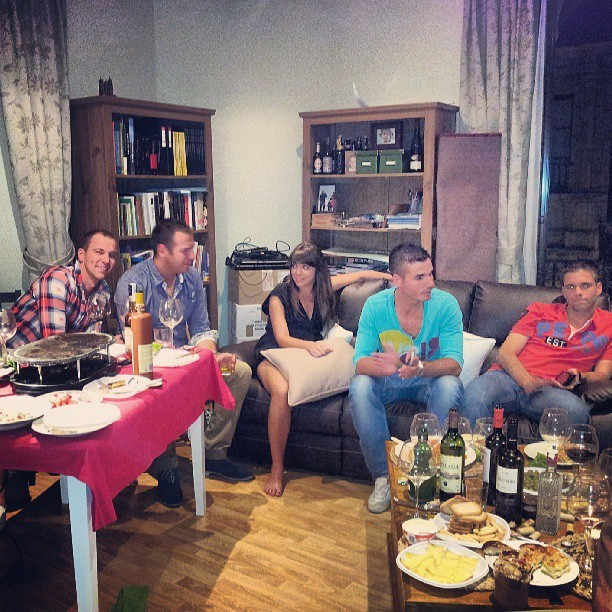}
        \caption*{Input Image (Social Gathering)}
    \end{minipage}
    \hfill
    \begin{minipage}{0.65\textwidth}
        \small
        \textbf{Base Generation (Original):}\\
        \textit{"This is a warm, candid photograph of a group of five friends gathered in a cozy living room for a dinner party. The scene is filled with the relaxed atmosphere of a social gathering. - \textbf{The People:} There are five individuals in the image, three men and two women... On the left, a man in a red and blue plaid shirt is smiling..."}
        
        \vspace{1.5em}
        
        \textbf{Steered Generation (Target Category: Wild Animal):}\\
        \textit{"The image captures a group of five animals in a natural setting, likely a savanna or grassland. The scene is dominated by a large herd of elephants, their massive bodies and long trunks moving gracefully across the open plain. Their gray, wrinkled skin is a striking contrast to the golden-brown grass that covers the landscape..."}
    \end{minipage}
    
    \vspace{1em}
    \caption{Cross-category steering in Qwen3-VL-30B at Layer 33. Despite the original image containing five humans in a domestic setting, the intervention on wild-animal-aligned neurons re-directs the model's description from a domestic gathering to a savanna scene with a herd of elephants. Note that the model fabricates environmental details (e.g., grassland) absent from the input, illustrating that strong steering can co-induce hallucination---a limitation discussed in Section ``Limitations and Future Work''.}
    \label{fig:steering_qwen_wild}
\end{figure*}

As shown in Figure~\ref{fig:steering_qwen_wild}, this case illustrates the observed steering effect on the MoE-based \textbf{Qwen3-VL-30B} model at Layer 33. By amplifying the \textit{wild animal} sparse circuits, the generated caption shifts an indoor social scene toward a savanna-like description featuring a herd of elephants.

\end{document}